\documentclass[11pt,a4paper]{article}
\usepackage[hyperref]{naaclhlt2018}
\usepackage{times}
\usepackage{latexsym}
\usepackage{graphicx}
\usepackage{mathtools}
\usepackage{tabularx}
\usepackage{amsmath}
\usepackage{booktabs} % For formal tables
\usepackage{url}
\usepackage{csquotes}
\usepackage{array}
\usepackage{colortbl}
\usepackage{ctable}
\newcolumntype{P}[1]{>{\centering\arraybackslash}p{#1}}
\newcolumntype{M}[1]{>{\centering\arraybackslash}m{#1}}
\usepackage{epstopdf}
\usepackage{hyperref}
\usepackage{xstring}
\usepackage{url}

\aclfinalcopy % Uncomment this line for the final submission
\newcommand\BibTeX{B{\sc ib}\TeX}

\title{Detecting Reliable Novel Word Senses: \\
A Network-Centric Approach}

\author{Abhik Jana \\
  IIT Kharagpur \\
  Kharagpur, India \\
  {\tt abhikjana1@gmail.com} \\\And
  Animesh Mukherjee \\
IIT Kharagpur \\
  Kharagpur, India \\
  {\tt animeshm@gmail.com}\\\And
  Pawan Goyal \\
IIT Kharagpur \\
  Kharagpur, India \\
  {\tt pawang@cse.iitkgp.ac.in} \\}

\date{}

\begin{document}
%\blindtext
\maketitle
\begin{abstract}
In this era of Big Data, due to expeditious exchange of information on the web, words are being used to denote newer meanings, causing linguistic shift. With the recent availability of large  amounts of  digitized texts, an automated analysis of the evolution of language has become possible. 
Our study mainly focuses on improving the detection of new word senses.  %Thus, novel sense detection becomes a crucial and challenging task in order to build any natural language processing application which depends on the efficient semantic representation of words. 
This paper presents a unique proposal based on network features to improve the precision of new word sense detection. For a candidate word where a new sense (birth) has been detected by comparing the sense clusters induced at two different time points, we further compare the network properties of the subgraphs induced from novel sense cluster across these two time points. Using the mean fractional change in edge density, structural similarity and average path length as features in an SVM classifier, manual evaluation gives precision values of 0.86 and 0.74 for the task of new sense detection, when tested on 2 distinct time-point pairs, in comparison to the precision values in the range of 0.23-0.32, when the proposed scheme is not used. The outlined method can therefore be used as a new post-hoc step to improve the precision of novel word sense detection in a robust and reliable way where the underlying framework uses a graph structure. Another important observation is that even though our proposal is a post-hoc step, it can be used in isolation and that itself results in a very decent performance achieving a precision of 0.54-0.62. Finally, we show that our method is able to detect the well-known historical shifts in 80\% cases.
\end{abstract}

\section{Introduction}
%%%\vspace{-0.1cm}
How do words develop new senses? How does one characterize semantic change? Is it possible to develop algorithms to track semantic change by comparing historical data at scale? In order to extract meaningful insights from these data, a very important step is to understand the contextual sense of a word, e.g., does the word `bass' in a particular context refer to fish or is it related to music?

Most data-driven approaches so far have been limited to either word sense induction where the the goal is to automatically induce different senses of a given word in an unsupervised clustering setting, or word sense disambiguation where a fixed sense inventory is assumed to exist, and the senses of a given word are disambiguated relative to the sense inventory. However in both these tasks, the assumption is that the number of senses that a word has, is static, and also the senses exist in the sense inventory to compare with. They attempt to detect or induce one of these senses depending on the context. However, natural language is dynamic, constantly evolving as per the users' needs which leads to change of word meanings over time. For example, by late 20$^\textrm{th}$ century, the word `float' has come up with the `data type' sense whereas the word `hot' has started corresponding to the `attractive personality' sense.  

\subsection{Recent advancements}
Recently, with the arrival of large-scale collections of historic texts and online libraries such as Google books, a new paradigm has been added to this research area, whereby the prime interest is in identifying the temporal scope of a sense~\cite{gulordava2011distributional,jatowt2014framework,tahmasebi2011towards,lau2014learning} which, in turn, can give further insights to the phenomenon of language evolution. Some recent attempts~\cite{kulkarni2015statistically,hamilton-leskovec-jurafsky:2016:P16-1,hamilton-leskovec-jurafsky:2016:EMNLP2016,eger-mehler-2016,TACL796} also have been made to model the dynamics of language in terms of word senses.

One of the studies in this area has been presented by Mitra {\sl et al.}~\shortcite{mitra2014s} where the authors show that at earlier times, the sense of the word `sick' was mostly associated to some form of illness; however, over the years, a new sense associating the same word to something that is `cool' or `crazy' has emerged. \iffalse In their work, authors consider multiple time points and not only detect new senses (i.e., `birth'), but also identify cases where (i) two senses become indistinguishable (`join'), or (ii) one sense splits into multiple senses, or (iii) a sense falls out of the vocabulary (`death').\fi Their study is based on a unique network representation of the corpus called a distributional thesauri (DT) network built using Google books syntactic n-grams. They have used unsupervised clustering techniques to induce a sense of a word and then compared the induced senses of two time periods to get the new sense for a particular target word.

\subsection{Limitations of the recent approaches} While Mitra {\sl et al.}~\shortcite{mitra2014s} reported a precision close to 0.6 over a random sample of 49 words, we take another random sample of 100 words separately and repeat manual evaluation.
When we extract the novel senses by comparing the DTs from 1909-1953 and 2002-2005, the precision obtained for these 100 words is as low as 0.32. Similarly if we extract the novel senses comparing the DTs of 1909-1953 with 2006-2008, the precision stands at 0.23.
We then explore another unsupervised approach presented in Lau {\sl et al.}~\shortcite{lau2014learning} over the same Google books corpus\footnote{\url{http://commondatastorage.googleapis.com/books/syntactic-ngrams/index.html}, we use `triarcs' dataset from `English All'}, apply topic modeling for sense induction and directly adapt their similarity measure to get the new senses. Using a set intersecting with the 100 random samples for Mitra {\sl et al.}~\shortcite{mitra2014s}, we obtain the precision values of 0.21 and 0.28, respectively. Clearly, none of the precision values are good enough for reliable novel sense detection. This motivates us to devise a new approach to improve the precision of the existing approaches. Further, being inspired by the recent works of applying
complex network theory in NLP applications like co-hyponymy detection~\cite{JANA18.78}, evaluating machine generated summaries~\cite{pardo2006using}, detection of ambiguity in a text~\cite{dorow2004using}, etc. we opt for a solution using complex network measures.  

\subsection{Our proposal and the encouraging results}
We propose a method based on the network features to reduce the number of false positives and thereby, increase the overall precision of the method proposed by Mitra {\sl et al.}~\shortcite{mitra2014s}. In particular, if a target word qualifies as a `birth' as per their method, we construct two induced subgraphs of those words that form the cluster corresponding to this `birth' sense from the corresponding distributional thesauri (DT) networks of the two time points. Next we compare the following three network properties: (i) the edge density, (ii) the structural similarity and (iii) the average path length~\cite{wasserman1994social,turnu2012entropy} of the two induced subgraphs from the two time points. A remarkable observation is that although this is a small set of only three features, for the actual `birth' cases, each of them has a significantly different value for the later time point and are therefore very discriminative indicators. In fact, the features are so powerful that even a small set of training instances is sufficient for making highly accurate predictions. 

\noindent\textbf{Results:} Manual evaluation of the results by 3 evaluators shows that this classification achieves an overall precision of 0.86 and 0.74 for the two time point pairs over the same set of samples, in contrast with the precision values of 0.32 and 0.23 by the original method. \if{0} Our approach also outperforms the case when a candidate sense is considered as `birth' only if both Mitra's and Lau's approaches flag it as `birth' and the reported `birth' clusters overlap. The proposed method can therefore be used as a novel post-hoc step to improve the reliability and robustness of novel word sense detection.\fi Note that we would like to stress here that an improvement of \textbf{more than double} in the precision of novel sense detection that we achieve has the potential to be the new stepping stone in many NLP and IR applications that are sensitive to novel senses of a word.

\subsection{Detecting known shifts} 
Further we also investigate the robustness of our approach by analyzing the ability to capture known historical shifts in meaning. Preparing  a list of words that have been suggested by different prior works as having undergone sense change, we see that 80\% of those words get detected by our approach. We believe that the ability to detect such diachronic shifts in data can significantly enhance various standard studies in natural language evolution and change.

\subsection{Impact}
We stress that our work could have strong repercussions in historical linguistics~\cite{bamman2011measuring}. Besides, lexicography is also expensive; compiling, editing and updating sense inventory entries frequently remains cumbersome and labor-intensive. Time specific knowledge would make the word meaning representations more accurate.
A well constructed semantic representation of a word is useful for many natural language processing or information retrieval systems like machine translation, semantic search, disambiguation, Q\&A, etc. For semantic search, taking into account the newer senses of a word can increase the relevance of the query result. Similarly, a disambiguation engine informed with the newer senses of a word can increase the efficiency of disambiguation, and recognize senses uncovered by the inventory that would otherwise have to be wrongly assigned to covered senses. Above all, a system having the ability to perceive the novel sense of a word can help in automatic sense inventory update by taking into account the temporal scope of senses.
\section{Related work}
\label{sec:rw}
Our work broadly classifies under data-driven models of language dynamics. One of the first attempts in this area was made by Erk~\shortcite{erk2006unknown}, where the author tried to model this problem as an instance of outlier detection, using
a simple nearest neighbor-based approach. Gulordava and Baroni~\shortcite{gulordava2011distributional} study the change in the semantic orientation of words using Google book n-grams corpus from different time periods. In another work, Mihalcea {\sl et al.}~\shortcite{mihalcea2012word} attempted to quantify the changes in word usage over time. \if{0} and came up with the intuition that changes in usage frequency and word senses contribute
to these differences in usages.\fi Along similar lines, Jatowt and Duh~\shortcite{jatowt2014framework} used the Google n-grams corpus from two different time points and proposed a method to identify semantic change based on the distributional similarity between the word vectors. Tahmasebi {\sl et al.}~\shortcite{tahmasebi2011towards} attempted to track sense changes from a newspaper corpus containing articles between 1785 and 1985. \if{0}Even though the interest for automatic identification of new word senses has grown, the research has been limited by the availability of appropriate
evaluation resources.\fi Efforts have been made by Cook {\sl et al.}~\shortcite{cook2014novel} to prepare the largest corpus-based dataset of diachronic sense differences. Attempts have been made by Lau {\sl et al.}~\shortcite{lau2012word} where they first introduced their topic modeling based word sense induction
method to automatically detect words with emergent novel senses
and in a subsequent work, Lau {\sl et al.}~\shortcite{lau2014learning} extended this task by leveraging the concept of predominant sense.
The first computational approach to track and detect statistically significant linguistic shifts of words has been proposed by Kulkarni {\sl et al.}~\shortcite{kulkarni2015statistically}. Recently, Hamilton {\sl et al.}~\shortcite{hamilton-leskovec-jurafsky:2016:P16-1} proposed a method to quantify semantic change by evaluating word embeddings against known historical changes. In another work, Hamilton {\sl et al.}~\shortcite{hamilton-leskovec-jurafsky:2016:EMNLP2016} categorized the semantic change into two types and proposed different distributional measures to detect those types. Attempts have also been made to analyze time-series model of embedding vectors as well as time-indexed self-similarity graphs in order to hypothesize the linearity of semantic change by Eger {\sl et al.}~\shortcite{eger-mehler-2016}. A dynamic Bayesian model of diachronic meaning change has been proposed by Frermann {\sl et al.}~\shortcite{TACL796}. \if{0} where they have shown novel sense detection task as one of their applications.\fi Recently, researchers have also tried to investigate the reasons behind word sense evolution and have come up with computational models based on chaining~\cite{ramiro2018algorithms}. Researchers also attempt to apply dynamic word embeddings as well to detect language evolution~\cite{rudolph2018dynamic,yao2018dynamic}, analyze temporal word analogy~\cite{szymanski2017temporal}.  

We now describe the two baselines that are relevant for our work.

\noindent \textbf{Baseline 1: Mitra {\sl et al.}~\shortcite{mitra2014s}} The authors proposed an unsupervised method to identify word sense changes automatically for nouns. 
\begin{figure*}[!tbh]
\centering
\begin{minipage}[b]{0.35\textwidth}
\centering
\includegraphics[width=1\textwidth]{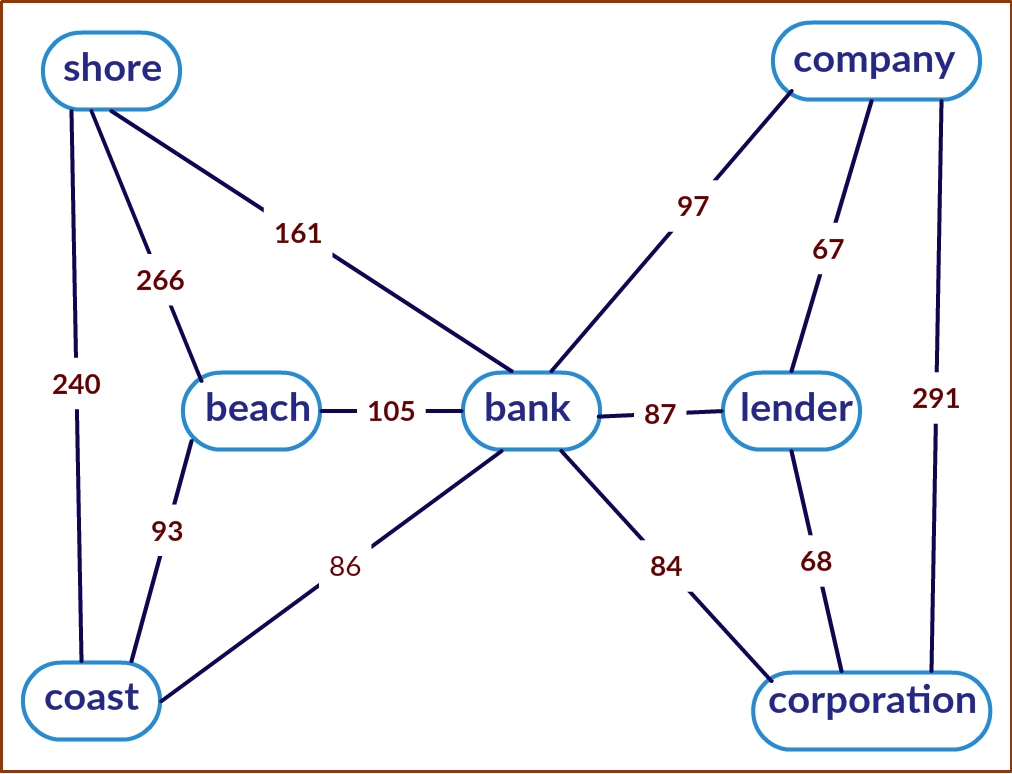}
\end{minipage}
\quad
\begin{minipage}[b]{0.46\textwidth}
\centering
\includegraphics[width=1\textwidth]{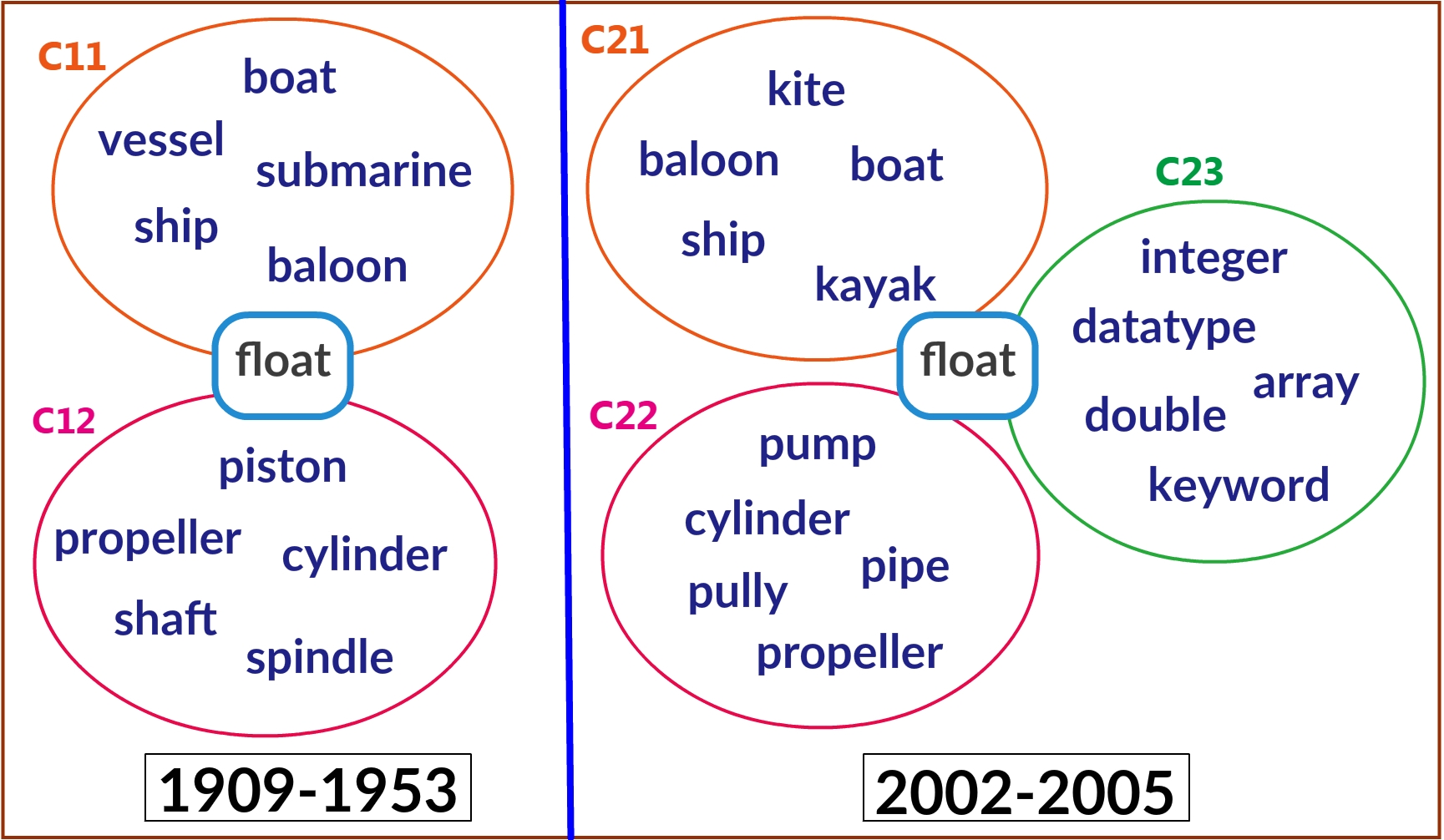}
\end{minipage}
%\vspace{-0.4cm}
\caption{Left image is a sample snapshot of the Distributional Thesaurus Network from the time period 2002-2005 where each node represents a word and the weight of the edge is defined as the number of context features that these two words share in common. %Here the word `bank' with some top distributionally similar words and the connections among them are shown.
Right image shows Chinese Whisper clusters for the target word `float' extracted from Google books syntactic n-gram corpus of both the time periods (1909-1953 and 2002-2005). A new sense of the word `float' has emerged with the `programming' related new cluster (C23) in 2002-2005.
}
\label{fig:DT}
%%%\vspace{-0.5cm}
\end{figure*}

\if{0}
\begin{figure}[!t]
%%%\vspace{-2mm}  
\centering
    \includegraphics[width=0.4\textwidth]{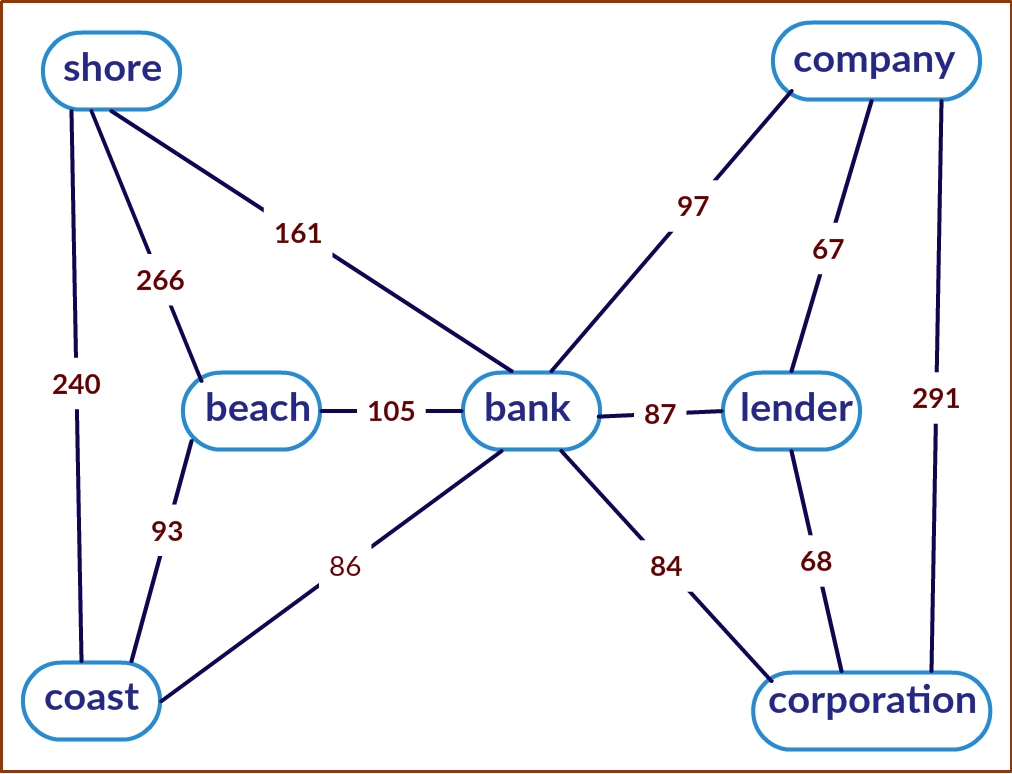}
 %%%\vspace{-0.1cm}  
\caption{A sample snapshot of Distributional Thesaurus Network from the time period 2002-2005 where each node represents a word and the weight of the edge is defined as the number of context features that these two words share in common. Here the word `bank' with some top distributionally similar words and the connections among them are shown.}
    \label{fig:DT}
%%%\vspace{-3mm}  
 %%%%%%\vspace{1mm}
\end{figure}
\fi

\noindent {\sl Datasets and graph construction:} The authors used the Google books corpus, consisting of texts from over 3.4 million digitized English books published between 1520 and 2008. The authors constructed distributional thesauri (DT) networks from the Google books syntactic n-grams data~\cite{goldberg2013dataset}. \iffalse The DT network contains, for each word, a list of words that are similar with respect to their bigram distribution~\cite{riedl2013scaling}.  In particular, they first extracted each word and a set of its context features like part-of-speech tag, neighbouring set of words, frequency etc. Next they calculated the lexicographer's mutual information (LMI)~\cite{kilgarriff2004itri}\footnote{$$
LMI(word,feature)=f(word,feature)*\log_2 (\nicefrac{f(word,feature)}{f(word)*f(feature)})$$, where $f()$ measures the frequency} between a word and its features and took the top 1000 ranked features for each word. \fi In the DT network, each word is a node and there is a weighted edge between a pair of words where the weight of the edge is defined as the number of features that these two words share in common. A snapshot of the DT is shown in leftmost image of Figure~\ref{fig:DT}. To study word sense changes over time, they divided the dataset across eight time periods; accordingly DT networks for each of these time periods were constructed separately. \iffalse The basic idea is that if a word undergoes a sense change, this can be detected by comparing its senses from two different time periods. The unsupervised method for inducing word senses in each time period is described below.\fi

\noindent {\sl Sense change detection:}
The Chinese Whispers algorithm~\cite{biemann2011structure} is used to produce a set of clusters for each target word by decomposing its neighbourhood in the DT network. The hypothesis is that different clusters signify different senses of a target word. The clusters for a target word `float' is shown in the right image of Figure~\ref{fig:DT}. The authors then compare the sense clusters extracted across two different time points to obtain the suitable signals of sense change. Specifically, for a candidate word $w$, a sense cluster in the later time period is called as a `birth' cluster if at least 80\% words of this cluster do not appear in any of the sense clusters from the previous time period. The authors then apply multi-stage filtering in order to obtain meaningful candidate words.

\if{0}
\subsubsection{\textbf{Unsupervised sense induction:}}
In order to get the induced sense clusters in an unsupervised way, Chinese Whispers algorithm~\cite{biemann2011structure} has been used. The algorithm produces a set of clusters for each target word by decomposing its neighbourhood in the DT network. The hypothesis is that different clusters signify different senses of a target word. The clusters for a target word `float' is shown in the left image of Figure~\ref{fig:DT}. The authors then compare the sense clusters extracted across two different time points to obtain the suitable signals of sense change. 

\if{0}
\begin{figure}[!tbh]
%%%\vspace{-0.2cm}

\centering

\includegraphics[width=0.75\textwidth,height=120pt]{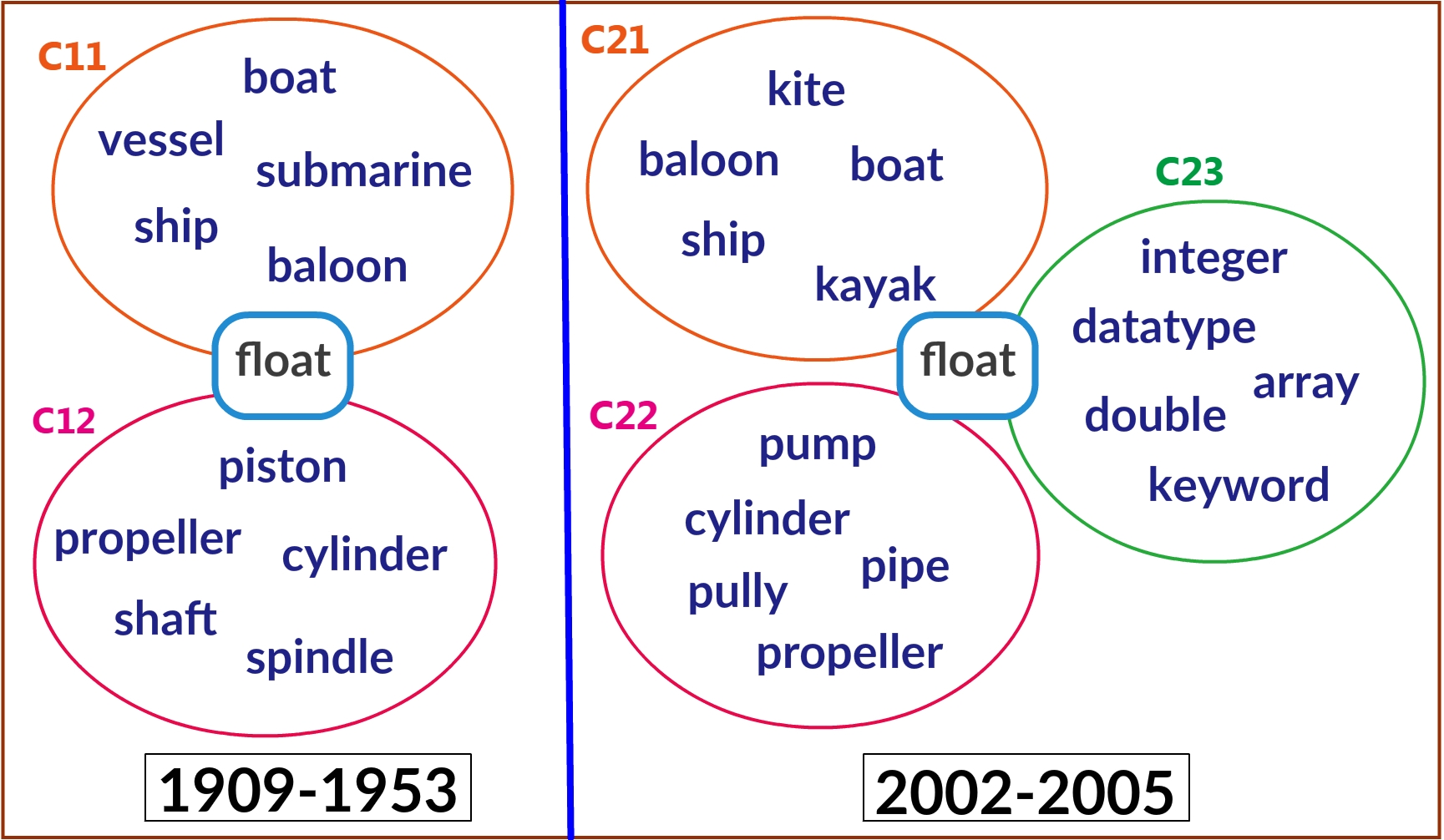}
%%%\vspace{-0.1cm}
\caption{\textbf{Chinese Whisper clusters for the target word `float' extracted from Google books syntactic n-gram corpus of both the time periods (1909-1953 and 2002-2005). A new sense of the word `float' has emerged with the `programming' related new cluster (C23) in 2002-2005.} %\worry{Make the words, circles, bounding lines everything more bold.}
}
\label{fig:gb}
%%%\vspace{-0.7cm}
\end{figure}
\fi

\subsubsection{\textbf{Sense change detection:}}
Let us assume that the Chinese Whispers algorithm is run over DTs corresponding to two different time periods, $tv_i$ and $tv_j$. Now, assume that for a given word $w$, the algorithm gives two different sets of clusters, $C_i$ and $C_j$, such that $m$ sense clusters are obtained in $tv_i$ and $n$ sense clusters are obtained in $tv_j$. Accordingly, let $C_i$ = {$s_{i1}$, $s_{i2}$, . . . , $s_{im}$} and 
$C_j$ = {$s_{j1}$, $s_{j2}$, . . . , $s_{jn}$}, where $s_{kz}$ denotes $z^{{th}}$ sense cluster for word $w$ during time interval $tv_k$. There are four types of sense changes that can happen for a target word -- \textit{split}, \textit{join}, \textit{birth} and \textit{death}. These sense changes are defined below.\\
{\bf split} A sense cluster $s_{iz}$ in $tv_i$ splits into two (or more) sense clusters, $s_{jp1}$ and $s_{jp2}$ in $tv_j$. \\
{\bf join}  Two sense clusters $s_{iz1}$ and $s_{iz2}$ in $tv_i$ join to make a single cluster $s_{jp}$ in $tv_j$. \\
{\bf birth} A new sense cluster $s_{jp}$ appears in $tv_j$, which was absent in $tv_i$. \\
{\bf death} A sense cluster $s_{iz}$ in $tv_i$ dies out and does not appear in $tv_j$. 

%\todo{Talk a bit about the thresholds used by the authors.}
%%%%%\vspace{-0.2cm}
A sense cluster is considered as `birth' if at least 80\% words of that cluster are novel, i.e., they do not appear in any of the clusters of old time points. 
For example, in Figure~\ref{fig:gb}, the programming related cluster of the target word `float' represents a `birth' sense. 
For split, each split cluster should have at least 30\% words of the source cluster and the total intersection of all the split clusters should be $>$ 80\%. For join and death, the same parameters are used with the interchange of the source and the target clusters. 

Note that, as our main focus is to detect novel sense of a word, we are concerned with only `birth' cases for our study.

\subsubsection{\textbf{Multi-stage filtering:}}

The authors then apply multi-stage filtering in order to obtain meaningful candidate words.\\
{\bf Stage 1} They apply Chinese Whisper three times over the two different time periods and take the intersection to output those clusters, which came up in all three runs. \\
{\bf Stage 2} As they focus only on nouns, they keep the candidate words tagged with \lq{NN}\rq or \lq{NNS}\rq. \\
{\bf Stage 3} They sort the target words based on their frequency counts and consider only the middle 60\% of the list which is the most informative part for this type of analysis. 
\fi
%We have considered only the `birth' of a new sense in our work. Specifically, for a candidate word $w$, a sense cluster in the later time period is called as a `birth' cluster if atleast 80\% words of this cluster do not appear in any of the sense clusters from the previous time period. 
\if{0}
For evaluation, the authors selected 49 candidate `birth' words from a total of 2789 candidate `birth' words while comparing 1909-1953 DT with the 2002-2005 DT. Using manual evaluation, 31 words were found to be true positives and 18 words were false positives. In our work we take these 49 candidate words and show that network features can be useful to discriminate the true positives from the false positives. 
\fi

\if{0}
The authors first construct distributional thesauri (DT) networks~\cite{riedl2013scaling} from the Google books syntactic n-grams data~\cite{goldberg2013dataset} 
at eight different time points. In a graph structure, the DT contains for each word a list of words that are similar with respect to their bigram distribution~\cite{riedl2013scaling}. 
Chinese Whispers algorithm~\cite{biemann2011structure} 
is then used to produce a set of clusters for each target word by decomposing its neighborhood in the DT network. The authors then compare the sense clusters extracted across two different time points to obtain the suitable signals of sense change. Specifically, for a candidate word $w$, a sense cluster in the later time period is called as a `birth' cluster if at least 80\% words of this cluster do not appear in any of the sense clusters from the previous time period. The authors then apply multi-stage filtering in order to obtain meaningful candidate words.
\fi
%%%%%%%\vspace{-0.2cm}

\noindent \textbf{Baseline 2: Lau {\sl et al.}~\shortcite{lau2014learning}:} The authors proposed an unsupervised approach based on topic modeling for sense induction, and showed novel sense identification as one of its applications. For a candidate word, Hierarchical Dirichlet Process~\cite{teh2006hierarchical} is run over a corpus to induce topics. The induced topics are represented as word multinomials,
and are expressed by the top-$N$
words in descending order of conditional probability. Each topic is represented as a sense of the target word. The words having highest probability in each topic represent the sense clusters. The authors treated the novel sense detection task as identifying those sense clusters, which did not align with any of the recorded senses in a sense repository. 
They used Jensen-Shannon (JS) divergence measure to compute alignment between a sense cluster and a synset. They computed JS divergence between the multinomial distribution
(over words) of the topic cluster and that of the synset, and converted the divergence value into a similarity score. Similarity between topic cluster $t_j$ and synset $s_i$ is defined as, 
%\begin{small}
\begin{equation}
%\vspace{-0.15cm}
sim(t_j,s_i) = 1-JS(T\parallel S)
\label{similarity}
%\vspace{-0.15cm}
\end{equation}
%\end{small}
where $T$ and $S$ are the multinomial distributions over words for topic $t_j$ and synset $s_i$, respectively,
and $JS(X\parallel Y)$ is the Jensen-Shannon divergence for the distribution $X$ and $Y$. Since we define novel senses while comparing sense clusters across two time points, we use the same JS measure to detect novel sense of a target word. A sense cluster in the newer time period denotes a new sense (`birth') only if its maximum similarity with any of the clusters in older time period is below a threshold, which we have set to 0.35 based on empirical observation.\\
\section{Proposed Network-Centric Approach}
\label{sec:method}
\begin{figure*}[!tbh]
\centering
\begin{minipage}[b]{0.47\textwidth}
\centering
\includegraphics[width=1\textwidth]{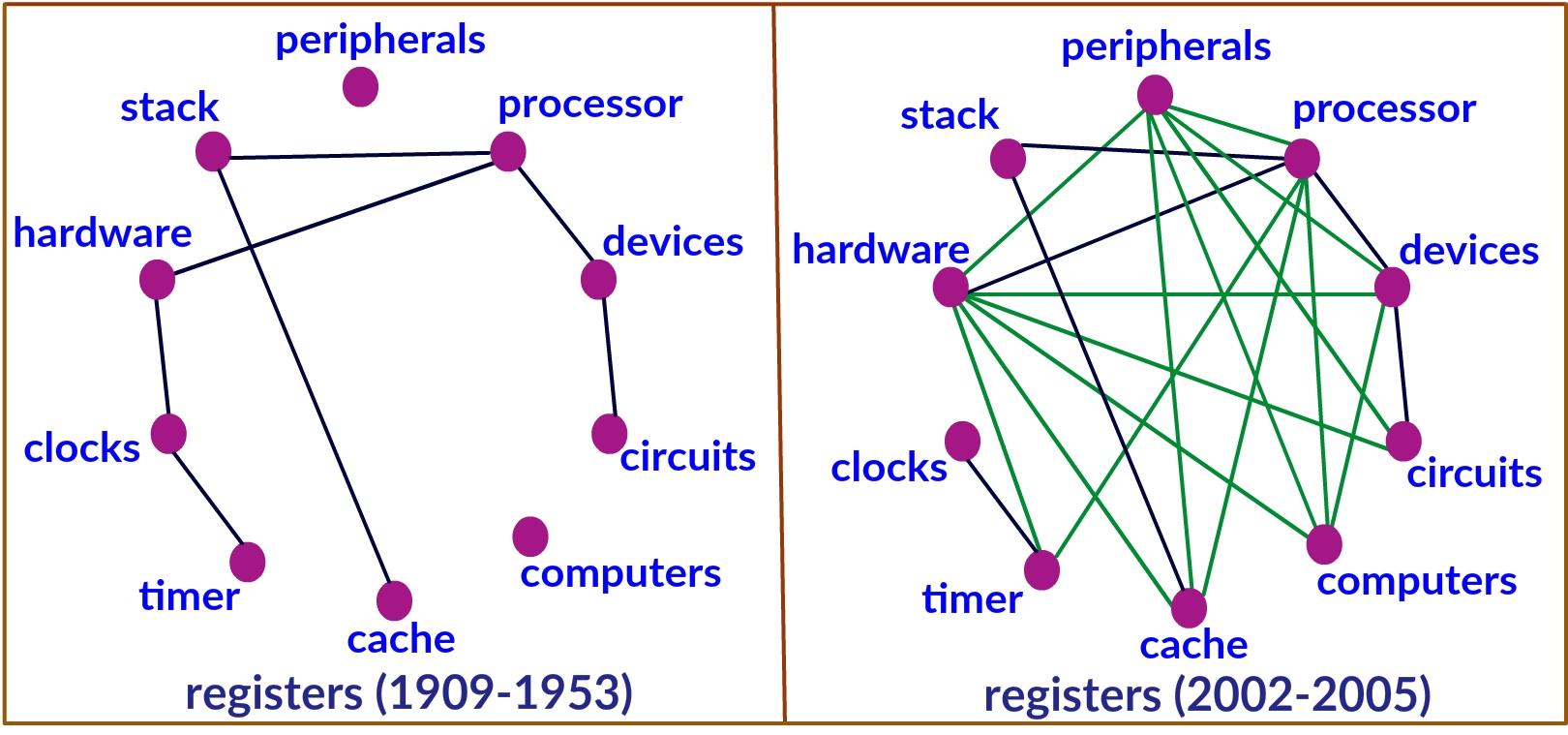}
\end{minipage}
\quad
\begin{minipage}[b]{0.47\textwidth}
\centering
\includegraphics[width=1\textwidth]{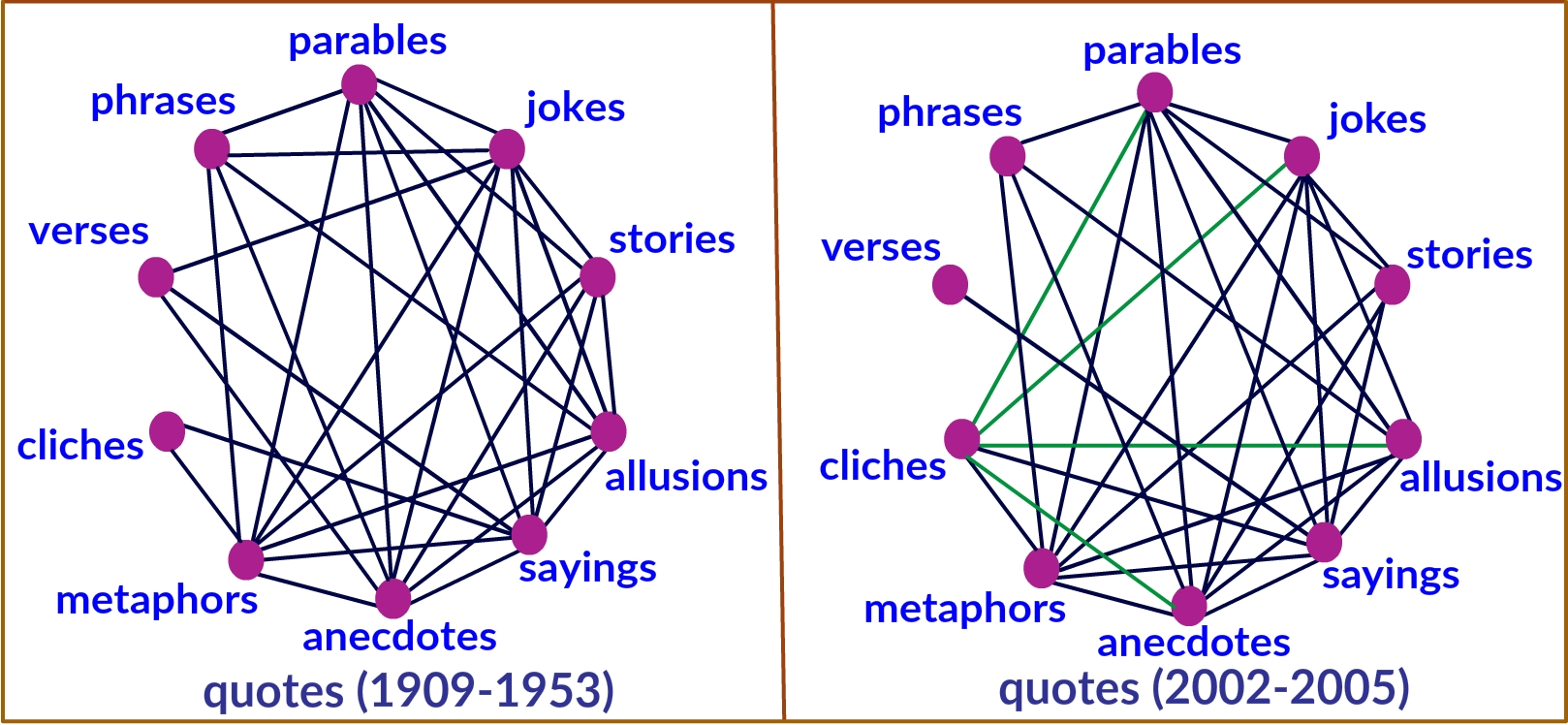}
\end{minipage}
%\vspace{-0.4cm}
\caption{Induced subgraphs of the `birth' clusters of `registers' and `quotes' for the two time periods (1909-1953 and 2002-2005). It shows that edge connections among the neighbours of `registers' have increased significantly over time which leads to emergence of `technology' related sense of `registers' whereas the connections among the neighbours of `quotes' are almost same over time, indicating non-emergence of any novel sense.}
\label{fig:induced}
%\vspace{-0.4cm}
\end{figure*}

Mitra {\sl et al.}~\shortcite{mitra2014s} selected 49 candidate `birth' words from a total of 2789 candidate `birth' words while comparing 1909-1953 DT with the 2002-2005 DT for manual evaluation; 31 words were found to be true positives and 18 words were false positives. We first study these 49 candidate `birth' words and show that network features can be useful to discriminate the true positives from the false positives. For each of these candidate words $w$, we take the `birth' cluster from 2002-2005, which is represented by a set of words $S$. According to our hypothesis, if the words in set $S$ together represent a new sense for $w$ in 2002-2005 which is not present in 1909-1953, the network connection among these words (including $w$) would be much more strong in the 2002-2005 DT than the 1909-1953 DT. The strength of this connection can be measured if we construct induced subgraphs of $S$ from the two DTs and measure the network properties of these subgraphs; the difference would be more prominent for the actual `birth' cases (true positives) than for the false `birth' signals (false positives). Note that by definition, the nodes in an induced subgraph from a DT will be the words in $S$ and there will be an edge between two words if and only if the edge exists in the original DT; we ignore the weight of the edge henceforth. Thus, the difference between the two subgraphs (one each from the older and newer DTs) will only be in the edge connections. 
Figure~\ref{fig:induced} takes one true positive (`register') and one false positive (`quotes') word from the set of 49 words and shows the induced subgraphs obtained by a subset of their `birth' clusters across the two time points. We can clearly see that connections among the words in $S$ is much stronger in the newer DT than in the older one in the case of `registers', indicating the emergence of a new sense. In the case of `quotes', however, the connections are not very different across the two time periods. We choose three \textit{cohesion indicating} network properties, (i) the edge density, (ii) the structural similarity and (iii) the average path length, to capture this change. 

Let $S=\{w_1,w_2,\ldots,w_n\}$ be the `birth' cluster for $w$. Once we construct a graph induced by $S$ from the DT, these network properties are measured as follows:

\noindent{\bf Edge Density (ED):} ED is given by
%\begin{small}
\begin{equation}
ED = N_a/N_p
\label{Edge_Density}
\end{equation}
%\end{small}
where $N_a$ denotes the number of actual edges between $w_1,w_2,\ldots,w_n$ and $N_p$ denotes the maximum possible edges between these, i.e., $\frac{n(n-1)}{2}$.
 
\noindent{\bf Structural Similarity (SS):} For each pair of words $(w_i,w_j)$ in the cluster $S$, the structural similarity $SS(w_i,w_j)$ is computed as:
%%%%%\vspace{-0.2cm}	 
%\begin{small}
\begin{equation}
%%%%%%\vspace{-0.2cm}
SS(w_i,w_j) = \frac{N_c}{\sqrt{deg(w_i)*deg(w_j)}}
\label{Structural_Similarity}
\end{equation}
%\end{small}
	where $N_c$ denotes the number of common neighbors of $w_i$ and $w_j$ in the induced graph and $deg(w_k)$ denotes the degree of $w_k$ in the induced graph, for $k=i,j$. The average structural similarity for the cluster $S$ is computed by averaging the structural similarity of all the word pairs.
    
\noindent{\bf Average Path Length (APL):}  To compute average path length of $S$, we first find the shortest path length between $w$ and the words $w_i$, in the induced graph of $S$. Let $spl_i$ denote the shortest path distance from $w$ to $w_i$. The average path length is defined as:
%\begin{small}
\begin{equation}
APL = \sum_i spl_i / n
\label{Degree_Entropy}
%%%%%\vspace{-0.2cm}
\end{equation}
%\end{small}
	
where $n$ is the number of words in the cluster $S$.

\begin{table*}[!tbh]
%\vspace{-0.2cm}
\small
\caption{The network properties of the induced subgraphs of a true positive (`registers') and a false positive (`quotes') for the time periods 1909-1953 ($t_1$) and 2002-2005 ($t_2$).} %The fractional changes ($\Delta$) in network properties are significantly higher for `registers' compared to `quotes'.}
%\vspace{-0.4cm}
\begin{center}
    \begin{tabular}{|M{1.2cm}|M{1.1cm}|M{1.1cm}|M{1cm}|M{1cm}|M{1.4cm}|M{1.4cm}|M{5cm}|}
    \hline
\bf{Word} &	\bf{ED ($t_1$)} & \bf{ED ($t_2$)} & \bf{SS ($t_1$)} & \bf{SS ($t_2$)} & \bf{APL ($t_1$)} & \bf{APL ($t_2$)} & \bf{$\Delta$ (ED, SS, APL)}\\ \hline
\emph{registers} &	0.108	& 0.546 &  0.076	& 0.516	&	1.9	& 1 &\cellcolor{green} 4.045, 5.771, -0.9 \\\hline %651.111 \\ \hline
quotes 	& 0.858 &	0.833 & 0.835	& 0.622 & 1.72	& 1 & \cellcolor{red!20}-0.029, -0.255, -0.72 \\\hline %	& 2.439 \\ \hline
    \end{tabular}
\end{center}
\label{tab:reg_quote}
%\vspace{-0.2cm}
\end{table*}

\begin{table*}[!tbh]
%\vspace{-0.2cm}
\small
\caption{Mean values of the network properties of the induced subgraphs of 31 true positives and 18 false positives for the time periods 1909-1953 ($t_1$) and 2002-2005 ($t_2$). The mean fractional changes ($\Delta$) in network properties are significantly higher for the true positives (TP) as compared to the false positives (FP).} 
%which indicates that the words emerged with new senses have undergone drastic change in network connections in their neighbourhood over time.}
%\vspace{-0.4cm}
\begin{center}
    \begin{tabular}{|M{1.2cm}|M{1.1cm}|M{1.1cm}|M{1cm}|M{1cm}|M{1.4cm}|M{1.4cm}|M{5cm}|}
    \hline
\bf{Word} &	\bf{ED ($t_1$)} & \bf{ED ($t_2$)} & \bf{SS ($t_1$)} & \bf{SS ($t_2$)} & \bf{APL ($t_1$)} & \bf{APL ($t_2$)} & \bf{$\Delta$ (ED, SS, APL)}\\ \hline
\emph{TP} &	0.34	& 0.772 &  0.311	& 0.647	&	1.941	& 1 &\cellcolor{green} 2.388, 4.654, -0.941\\\hline %651.111 \\ \hline
FP	& 0.576 & 0.828 & 0.574	& 0.681 & 1.828	& 1 &\cellcolor{red!20} 0.747, 0.507, -0.828 \\\hline %	& 2.439 \\ \hline
    \end{tabular}
\end{center}
\label{tab:mean_properties}
%\vspace{-0.5cm}
\end{table*}
Table \ref{tab:reg_quote} notes the values obtained for these network properties for the induced subgraphs of the reported `birth' clusters for `registers' and `quotes' across the two time periods. The fractional changes observed for the three network properties show a clear demarcation between the two cases. Fractional change ($\Delta$) of any network measure $P$ is defined as,
\begin{equation}
 \Delta(P)= (P(t_2)-P(t_1)) / P(t_1)
\label{fc}
\end{equation}

where $t_1$ and $t_2$ are old and new timeperiods respectively. The change observed for the `birth' cluster of `registers' is significantly higher than that in `quotes'\footnote{As we have taken the `birth' clusters from new time period ($t_2$), the words in the clusters are the direct neighbors of the target word always resulting in average path length of 1 in $t_2$}. 

We now compute these parameter values for all the 49 candidate cases. The mean values obtained for the true positives (TP) and false positives (FP) are shown in Table~\ref{tab:mean_properties}. The findings are consistent with those obtained for `registers' and `quotes'.

We, therefore, use the fractional changes in the three network properties over time as three features to classify the remaining candidate `birth' words into true positives (actual `birth') or false positives (false `birth'). 
\section{Experimental results}
\label{sec:ex}
%%\vspace{-0.2cm}
For experimental evaluation, we start with the `birth' cases reported by Mitra {\sl et al.}~\shortcite{mitra2014s} -- 2740 cases (after removing the 49 used in training) for 1909-1953 -- 2002-2005 ($T_1$) and 2468 cases for 1909-1953 -- 2006-2008 ($T_2$). We run Lau {\sl et al.}~\shortcite{lau2014learning} over these birth cases to detect `novel' sense as per their algorithm. Separately, we also apply the proposed SVM classification model as a filtering step to obtain `filtered birth' cases. This helps in designing a {\sl comparative evaluation} of these algorithms as follows. From both the time point pairs ($T_1$ and $T_2$), we take 100 random samples from the birth cases reported by Mitra {\sl et al.}~\shortcite{mitra2014s} and get these manually evaluated. For the same 100 random samples, we now use the outputs of Lau {\sl et al.}~\shortcite{lau2014learning} and the proposed approach, and estimate the precision as well as recall of these. 

To further evaluate the proposed algorithm, we perform two more evaluations. First, we take 60 random samples from each time point pair for computing precision of the `filtered birth' cases. Secondly, we also take 100 random samples for each time point pair for computing precision of our approach independently of Mitra {\sl et al.}~\shortcite{mitra2014s}, i.e., the proposed approach is not informed of the `birth' cluster reported by  Mitra {\sl et al.}~\shortcite{mitra2014s}, instead all the clusters in old and new time point are shown.

We perform all the evaluations manually and each of the candidate word is judged by 3 evaluators. These evaluators are graduate/post-graduate students, having good background in Natural Language Processing. They are unaware of each other, making the annotation process completely blind and independent. Evaluators are shown the detected `birth' cluster from the newer time period and all the clusters from the older time period. They are asked to make a binary judgment as to whether the `birth' cluster indicates a new sense of the candidate word, which is not present in any of the sense clusters of the previous time point\footnote{An anonymized sample evaluation page can be seen here: \url{https://kwiksurveys.com/s/7TfSoYF2}}. Majority decision is taken in case of disagreement. In total, we evaluate the system for a set of as large as \textbf{520} words\footnote{100+100+60, per time point pair ($T_1$ and $T_2$)} which we believe is significant given the tedious manual judgment involved.

In this process of manual annotation, we obtain an inter-annotator agreement (Fleiss' kappa~\cite{fleiss1971measuring}) of $\bf0.745$, which is {\em 	substantial}~\cite{viera2005understanding}.
Table~\ref{tab:examples} shows three example words from $T_1$, their `birth' clusters as reported in Mitra {\sl et al.}~\shortcite{mitra2014s} and the manual evaluation result. The first two cases belong to computer related sense of `searches' and `logging', which were absent from time point 1909-1953. On the other hand, the `birth' cluster of `pesticide' represents an old sense which was also present in 1909-1953. Similarly Table~\ref{tab:examples_lau} shows manual evaluations results for 3 example cases, along with their novel sense as captured by Lau {\sl et al.}~\shortcite{lau2014learning}.

\begin{table}[!tbh]
%\vspace{-0.2cm}
\small
\caption{Example `birth' clusters reported in Mitra {\sl et al.}~\shortcite{mitra2014s} and manual evaluation.}
%For each word, `birth' cluster represents the words in the Chinese Whisper cluster which is flagged as new sense by the model.}
%\vspace{-0.3cm}

\begin{center}
    \begin{tabular}{ |M{1cm}|M{2.7cm}|M{2.8cm}|}
    \hline
    \centering \bf{Word} & \centering \bf{`birth' cluster} & \bf{Manual Evaluation}\\ \hline
    \centering \emph{searches} & folders, templates, syntax, formats, $\ldots$ & \textbf{Yes}, technology related sense \\ \hline
    \centering \emph{logging} & server, console, security, service, $\ldots$ & \textbf{Yes}, technology related sense \\ \hline
     \centering pesticide & fertilizer, sediment, waste,  $\ldots$ & \textbf{No}  \\ \hline
    
    \end{tabular}
\end{center}
\label{tab:examples}
%%\vspace{-0.4cm}
\end{table}

\begin{table}[!tbh]
\small

\caption{Example novel senses as per Lau {\sl et al.}~\shortcite{lau2014learning} and manual evaluation.}
%`Novel sense' represents the list of words which indicates a new sense as per Lau {\sl et al.}~\shortcite{lau2014learning}.}
%\vspace{-0.4cm}

\begin{center}
    \begin{tabular}{|M{1.4cm}|M{2.3cm}|M{2.8cm}|}
    \hline
  \centering \bf{Word} & \centering \bf{Novel sense} & \bf{Manual Evaluation} \\ \hline
    \centering \emph{stereo} & system, player, computer, $\ldots$ & \textbf{Yes}, technology related sense \\ \hline
    \centering \emph{mailbox} & email, pages, postal, $\ldots$ & \textbf{Yes}, technology related sense \\ \hline
 \centering acidification & acidosis, renal, distal, urinary,  
 $\ldots$ & \bf{No} \\ \hline
    
    \end{tabular}
\end{center}
\label{tab:examples_lau}
%\vspace{-0.3cm}
\end{table}

\noindent \textbf{Comparative evaluation: }
Only 32 and 23 words out of the 100 random samples from two time point pairs are evaluated to be actual `birth's, respectively, thus giving precision scores of 0.32 and 0.23 for Mitra {\sl et al.}~\shortcite{mitra2014s}. 
Evaluation results for the same set of random samples after applying the approach outlined in~\citet{lau2014learning} are presented in Table~\ref{tab:result_summary_lau_approach}. Since the reported novel sense cluster can in principle be different from the `birth' sense reported by the method of~\citet{mitra2014s} for the same word, we get the novel sense cases manually evaluated by 3 annotators (42 and 28 cases for the two time periods, respectively). Note that for these 100 random samples (that are all marked `true' by~\citet{mitra2014s}), it is possible to find an upper bound on the recall of~\citet{lau2014learning}'s approach automatically. While the low recall might be justified because this is a different approach, even the precision is found to be in the same range as that of~\citet{mitra2014s}.

Table~\ref{tab:result_summary_our_approach} presents the evaluation results for the same set of 100 random samples after using the proposed SVM filtering. We see that the filtering using SVM classification improves the precision for both the time point pairs ($T_1$ and $T_2$) significantly, boosting it from the range of 0.23-0.32 to 0.74-0.86. Note that, as per our calculations, indeed the recall of ~\citet{mitra2014s} would be 100\% (as we are taking random samples for annotation from the set of reported `birth' cases by ~\citet{mitra2014s} only). Even then ~\citet{mitra2014s}'s F-measure ranges from 0.37-0.48 while ours is 0.67-0.68. 
Table~\ref{tab:examples_FN} represents some of the examples which were declared as `birth' by Mitra {\sl et al.}~\shortcite{mitra2014s} but SVM filtering correctly flagged them as `false birth'. The feature values in the third column clearly show that the network around the words in the detected `birth' cluster did not change much and therefore, the SVM approach could correctly flag these. Considering the small training set, the results are highly encouraging. We also obtain decent recall values for the two time point pairs, giving an overall F-measure of 0.67-0.68.

\begin{table}[!tbh]
\small
%\vspace{-0.2cm}

\footnotesize
\caption{Evaluation of the approach presented in Lau {\sl et al.}~\shortcite{lau2014learning} with accuracy for 100 random samples.}
%%\vspace{-0.4cm}

\begin{center}
    \begin{tabular}{|M{0.8cm}|M{1.6cm}|M{1.1cm}|M{0.7cm}|M{1.3cm}|}
    \hline
Time- & \multicolumn{4}{c|}{\bf{Lau {\sl et al.}~\shortcite{lau2014learning}}} \\\cline{2-5}
point & \bf{\# Novel senses} & \bf{Precision} & \bf{Recall} & \bf{F-measure} \\\hline
\rowcolor{red!20}$T_1$ & 1189 & 0.21 & 0.28 & 0.24 \\\hline
\rowcolor{red!20}$T_2$ & 787 & 0.28 & 0.35 & 0.31  \\\hline

    \end{tabular}
\end{center}
\label{tab:result_summary_lau_approach}
%\vspace{-0.5cm}
\end{table}

\begin{table}[!tbh]
%%\vspace{-0.1cm}
\small
\footnotesize
\caption{Evaluation of the SVM-based filtering with accuracy reported for 100 random samples.}
%%\vspace{-0.4cm}
\begin{center}
    \begin{tabular}{|M{0.8cm}|M{1.4cm}|M{1.1cm}|M{0.7cm}|M{1.3cm}|}
    \hline
Time- & \multicolumn{4}{c|}{\bf{SVM filtering}} \\\cline{2-5}
point & \bf{\# birth cases} & \bf{Precision} & \bf{Recall} & \bf{F-measure} \\\hline
\rowcolor{green}$T_1$ & 318 & \textbf{0.86} & 0.56 & 0.68 \\\hline
\rowcolor{green}$T_2$ & 329 & \textbf{0.74} & 0.61 & 0.67  \\\hline

    \end{tabular}
\end{center}
\label{tab:result_summary_our_approach}
%\vspace{-0.3cm}
\end{table}

\begin{table}[!tbh]
%\vspace{-0.1cm}
\footnotesize
\caption{Example cases, which Mitra {\sl et al.}~\shortcite{mitra2014s} declared as true `birth' but SVM filtering correctly filtered}
%%\vspace{-0.3cm}

\begin{center}
    \begin{tabular}{|M{1cm}|M{3.5cm}|M{1.6cm}|}
    \hline
\centering \bf{Word} & \centering \bf{`birth' cluster} & $\Delta$(\bf{ED, SS, APL})  \\ \hline
\centering guaranty & acknowledgement, presumption, kind,  $\ldots$ & 0.11, -0.07, -0.5 \\\hline 
\centering troll & shellfish, salmon, bait, trout, tuna,  $\ldots$ & -0.04, -0.18, -0.84\\\hline
\centering nightcap & supper, lunch, dinner, nap,  luncheon, $\ldots$ & 0.04, -0.17, -0.75\\ \hline
    \end{tabular}
\end{center}
\label{tab:examples_FN}
%\vspace{-0.5cm}
\end{table}

Further, we check if we can meaningfully combine the results reported by both the methods of Mitra {\sl et al.}~\shortcite{mitra2014s} and Lau {\sl et al.}~\shortcite{lau2014learning} for more accurate sense detection; and how does this compare with the SVM based filtering. Therefore, we filter the words, which are reported as `birth' by both these methods and the reported `birth' sense clusters have a non-zero overlap. Out of 2789 and 2468 cases reported as `birth' by the method of Mitra {\sl et al.}~\shortcite{mitra2014s}, we obtain 132 and 86 cases respectively as having an overlapping sense cluster with that obtained using Lau's method. Two such examples are shown in Table~\ref{tab:examples_mitra_lau}; both the senses look quite similar. Table~\ref{tab:intersection_with_mitra} shows the accuracy results obtained using this approach. Only 6 and 2 words out of those 100 samples were flagged as `birth' for the two time points $T_1$ and $T_2$ respectively. Thus, the recall is very poor. While the precision improves slightly for $T_1$ (4 out of 6 are correct), it is worse for $T_2$ (only 1 out of 6 words is correct). The results confirm that the proposed SVM classification approach works better than both the approaches, individually as well as combined. 
\begin{table}[!tbh]
%\vspace{-0.2cm}
\footnotesize
\caption{Examples cases, which Mitra {\sl et al.}~\shortcite{mitra2014s} declared as `birth' represent the same sense as obtained using Lau {\sl et al.}~\shortcite{lau2014learning} ($T_1$).}
%\vspace{-0.4cm}
\begin{center}
    \begin{tabular}{|M{1cm}|M{2.8cm}|M{2.8cm}|}
    \hline
\bf{Word} & \bf{`birth' cluster as reported inMitra {\sl et al.}~\shortcite{mitra2014s}} & \bf{Novel senses as obtained usingLau {\sl et al.}~\shortcite{lau2014learning}}  \\ \hline
burgers & rice, pizza, \underline{fries}, \underline{drinks}, \underline{entrees}, \underline{desserts} $\ldots$ & \underline{fries}, orders, \underline{drinks}, \underline{entrees}, \underline{desserts} $\ldots$ \\\hline 
semantic & \underline{syntactic}, analytic, \underline{pragmatic}, \underline{lexical}, metaphoric $\ldots$ & \underline{syntactic}, \underline{pragmatic}, \underline{lexical}, aspect, context $\ldots$ \\\hline
    \end{tabular}
\end{center}
\label{tab:examples_mitra_lau}
%\vspace{-0.4cm}
\end{table}

\begin{table}[!tbh]
%\vspace{-0.2cm}
\small
\footnotesize
\caption{Evaluation of the intersection set while taking gold standard annotation of Mitra {\sl et al.}~\shortcite{mitra2014s}.}
%\vspace{-0.4cm}

\begin{center}
    \begin{tabular}{|M{1.7cm}|M{1.3cm}|M{1.6cm}|M{1.5cm}|}
    \hline
%Time- & \multicolumn{3}{c|}{SVM } \\\cline{2-4}
\bf{time point} & \bf{Precision} & \bf{Recall} & \bf{F-measure} \\\hline
$T_1$ & 0.67 (4/6) & 0.13 (4/32) & 0.22 \\\hline
$T_2$ & 0.5 (1/2) & 0.043 (1/23) & 0.08  \\\hline

    \end{tabular}
\end{center}

\label{tab:intersection_with_mitra}
%\vspace{-0.3cm}
\end{table}

\noindent \textbf{Feature analysis:} We therefore move onto further feature analysis and error analysis of the proposed approach. To validate the usefulness of all the identified features, we perform feature leave-one-out experiments. The results for $T_1$ are presented in Table~\ref{tab:t1} and~\ref{tab:t2}. We see that F-measure drops as we leave out one of the features. While $\{ED, SS\}$ turns out to be the best for precision, $\{SS, APL\}$  gives the best recall. Table~\ref{tab:examples_l1out} provides three examples to illustrate the importance of using all the three features. For the word `newsweek', using $\{ED,APL\}$ 
and for the word `caring', using $\{ED, SS\}$ could not detect those as `birth'. Only when all the three features are used, these cases are correctly detected as `birth'. Edge density, on the other hand is very crucial for improving precision. For instance, when only $\{SS,APL\}$ are used, words like `moderators' are wrongly flagged as `true'. Such cases are filtered out when all the three features are used.
\begin{table}[!tbh]
%\vspace{-0.1cm}
\footnotesize
\caption{Feature leave-one-out results ($T_1$). 
%It shows the importance of taking all three features together which gives us the best performance for the time point 1909-1953 - 2002-2005 ($T_1$).
}

%\vspace{-0.1cm}
\begin{center}
    \begin{tabular}{|M{2.2cm}|M{1.1cm}|M{1cm}|M{1.6cm}|}
    \hline
    \bf{Features used} & \bf{Precision} & \bf{Recall} & \bf{F-measure}\\ \hline
\centering $\Delta$(ED, SS) & 0.85 & 0.53 & 0.65 \\ \hline
\centering $\Delta$(ED, APL) & 0.84 & 0.5 & 0.62 \\ \hline
\centering $\Delta$(SS, APL) & 0.81 & 0.56 & 0.66 \\ \hline
\centering $\Delta$(ED, SS, APL) & \cellcolor{green}\textbf{0.86} & 0.56 & \cellcolor{green}\textbf{0.68}  \\ \hline
    \end{tabular}
\end{center}
\label{tab:t1}
%%\vspace{-0.3cm}
\end{table}

\begin{table}[!tbh]
%\vspace{-0.1cm}
\footnotesize
\caption{Feature leave-one-out results ($T_2$). %It shows the importance of taking all three features together which gives us the best performance for the time point 1909-1953 - 2006-2008 ($T_2$).
}
%\vspace{-0.1cm}
\begin{center}
    \begin{tabular}{|M{2.2cm}|M{1.1cm}|M{1cm}|M{1.6cm}|}
    \hline
    \bf{Features used} & \bf{Precision} & \bf{Recall} & \bf{F-measure}\\ \hline
\centering $\Delta$(ED, SS) & 0.72 & 0.56 & 0.63 \\ \hline
\centering $\Delta$(ED, APL) & 0.73 & 0.6 & 0.66 \\ \hline
\centering $\Delta$(SS, APL) & 0.66 & 0.61 & 0.63 \\ \hline
\centering $\Delta$(ED, SS, APL) & \cellcolor{green}\textbf{0.74} & 0.61 & \cellcolor{green}\textbf{0.67} \\ \hline
    \end{tabular}
\end{center}
\label{tab:t2}
%%\vspace{-0.3cm}
\end{table}

\begin{table}[!tbh]
%%\vspace{-0.1cm}
\footnotesize
\caption{Example cases to show the utility of all the features ($T_1$). The true positive cases like `newsweek' and `caring' get successfully detected whereas `moderators' gets successfully detected as false positive if all the three features are considered together.}
%\vspace{-0.4cm}
\begin{center}
    \begin{tabular}{|M{1.5cm}|M{3cm}|M{2.2cm}|}
    \hline
\centering \bf{Word} & \centering \bf{`birth' cluster} & $\Delta$(\bf{ED, SS, APL}) \\ \hline
\centering \emph{newsweek} & probation, counseling, $\ldots$ & 0.82, 1.58, -1.3 \\ \hline 
\centering \emph{caring} & insightful, wise, benevolent, $\ldots$ & 0.2, 0.13, -2.21 \\ \hline 
\centering moderators & correlate, function, determinant, $\ldots$ & 0.56, 0.44, -1.78 \\\hline
    \end{tabular}
\end{center}
\label{tab:examples_l1out}
\end{table}

\noindent \textbf{Extensive evaluation of the proposed approach:} We first take 60 random samples each from the `birth' cases reported by the SVM filtering for the two time point pairs, $T_1$ (from 318 cases) and $T_2$ (from 329 cases). The precision values of this evaluation are found to be \textbf{0.87} (52/60) and \textbf{0.75} (45/60) respectively, quite consistent with those reported in Table~\ref{tab:result_summary_our_approach}. We did another experiment in order to estimate the performance of our model for detecting novel sense, independent of the method of Mitra {\sl et al.}~\shortcite{mitra2014s}. We take 100 random words from the two time point pairs ($T_1$ and $T_2$), along with all the induced clusters from the newer time period and run the proposed SVM filtering approach to flag the novel `birth' senses. According to our model, for $T_1$ and $T_2$ respectively, 16 and 15 words are flagged to be having novel sense achieving precision values of 0.54 and 0.62 on manual evaluation, which itself is quite decent. Note that, for some cases, multiple clusters of a single word have been flagged as novel senses and we observe that these clusters hold a similar sense.     

\noindent \textbf{Error analysis:} We further analyze the cases, which are labeled as `true birth' by the SVM but are evaluated as `false' by the human evaluators. We find that in most of such cases, the sense cluster reported as `birth' contained many new terms (and therefore, the network properties have undergone change) but the implied sense was already present in one of the previous clusters with 
{\sl very few common words} (and therefore, the new cluster contained $>80\%$ new words, and is being reported as `birth' in Mitra {\sl et al.}~\shortcite{mitra2014s}). Two such examples are given in Table~\ref{tab:error_analysis}. The split-join algorithm proposed in Mitra {\sl et al.}~\shortcite{mitra2014s} needs to be adapted for such cases.

\begin{table}[!tbh]
%\vspace{-0.2cm}
\small
\footnotesize
\caption{Example `false positives' after SVM filtering ($T_1$). These words are flagged `true birth' by SVM but manually evaluated as `false'.}% Old cluster from 1909-1953 represents a similar sense as the `birth' cluster from 2002-2005.}
%\vspace{-0.4cm}
\begin{center}
    \begin{tabular}{|M{1.8cm}|M{2.3cm}|M{2.4cm}|}
    \hline
\centering \bf{Word} & \centering \bf{`birth' cluster} & \bf{Old cluster} \\\hline
\centering aftercare & care, clinic, outpatient, $\ldots$ & treatment, therapy, hospitalization, $\ldots$ \\\hline
\centering electrophoresis & labeling, analysis, profiling, $\ldots$ & analysis, counting, procedure,  $\ldots$  \\\hline
\end{tabular}
\end{center}
\label{tab:error_analysis}
\end{table}  

We also analyze the `false positive' cases, which are labeled as `false birth' by the SVM filtering but are evaluated as `true' by the human evaluators. Two such examples are given in Table~\ref{tab:examples_TN}. By looking at the feature values of these cases, it is clear that the network structure of the induced subgraph is not changing much, yet they undergo sense change. The probable reason could be that the target word was not in the network of the induced subgraph in the old time point and enters into it in the new time point. Our SVM model is unable to detect this single node injection in a network so far. Handling these cases would be an immediate future step to improve the recall of the system. 

\begin{table}[!tbh]
\small
\footnotesize
\caption{Example cases, labeled by SVM as `false birth' but flagged as `true birth' by annotators ($T_1$). The fractional change of the network measures is very low, leading to erroneous classification by SVM.}

%\vspace{-0.4cm}
\begin{center}
    \begin{tabular}{|M{1.1cm}|M{3cm}|M{2.3cm}|}
    \hline
\centering \bf{Word} & \centering \bf{`birth' cluster} & $\Delta$(\bf{ED, SS, APL}) \\ \hline
\centering baseplate & flywheel, cylinder, bearings, $\ldots$ & 0.06, -0.08, -0.84 \\ \hline 
\centering grating & beam, signal, pulse, $\ldots$ & 0.2, -0.05, -0.88 \\\hline
    \end{tabular}
\label{tab:examples_TN}
    \end{center}
\end{table}

\section{Detection of known shifts}
So far, we have reported experiments on discovering novel senses from data and measured the accuracy of our method using manual evaluation. In this section, we evaluate the diachronic validity of our method on another task of detecting known shifts. We test, whether our proposed method is able to capture the known historical shifts in meaning. For this purpose, we create a reference list $L$ of \textbf{15} words that have been suggested by prior work~\cite{eger-mehler-2016,hamilton-leskovec-jurafsky:2016:P16-1,hamilton-leskovec-jurafsky:2016:EMNLP2016} as having undergone a linguistic change and emerging with a novel sense. Note that, we focus only on nouns that emerge with a novel sense between 1900 and 1990. The goal of this task is to find out the number of cases, our method is able to detect as novel sense from the list $L$, which in turn would prove the robustness of our method.\\
\noindent {\bf Data:} Consistent with the prior work, we use the Corpus of Historical American (COHA)\footnote{\url{https://corpus.byu.edu/coha}}. COHA corpus is carefully created to be genre balanced and is a well constructed prototype of American English over 200 years, from the time period 1810 to 2000. We extract the raw text data of two time slices: 1880-1900 and 1990-2000 for our experiment.\\
\begin{figure}[!tbh]
%\vspace{-0.4cm}
\centering
\includegraphics[width=0.47\textwidth]{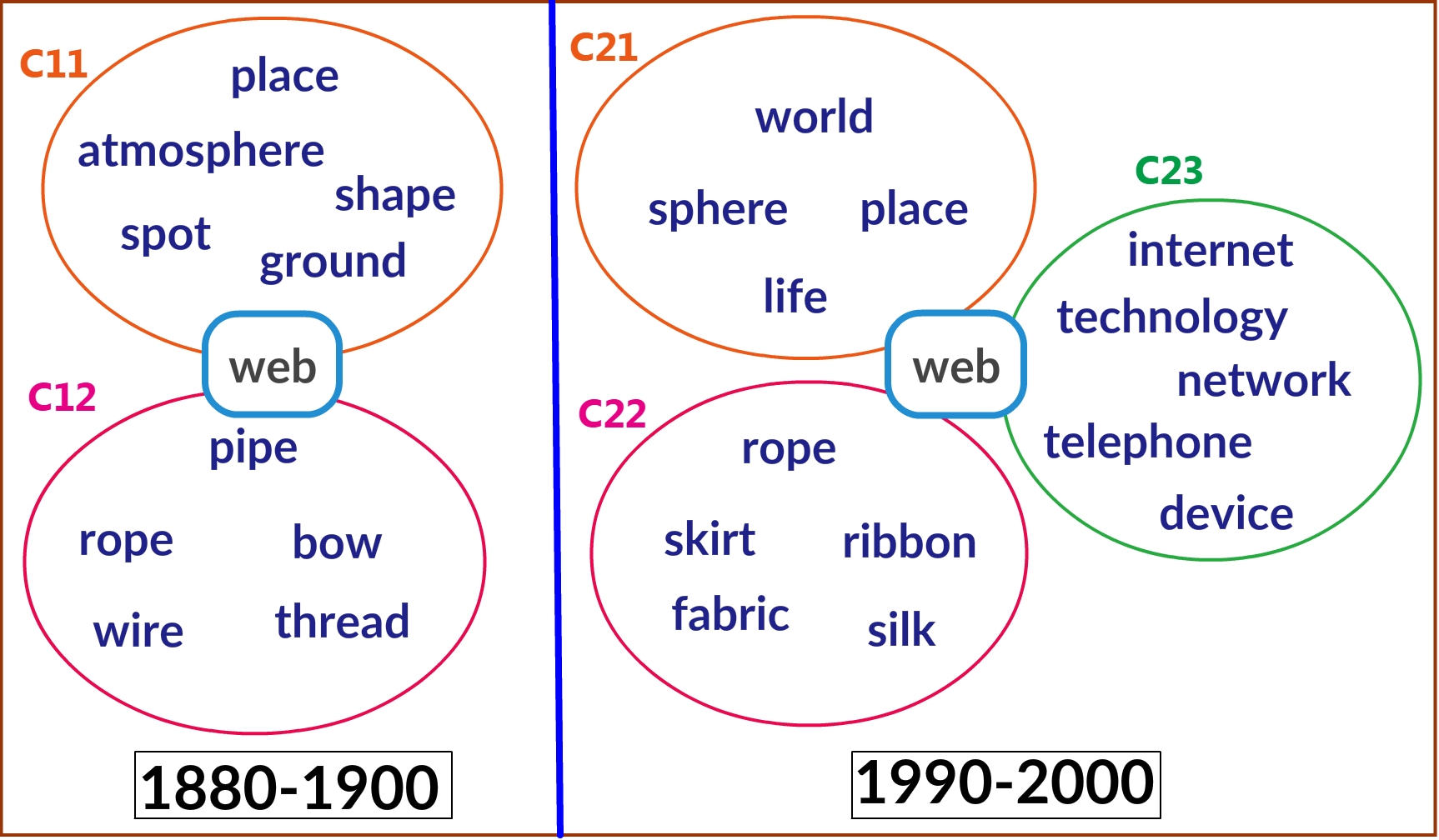}
%\vspace{-0.4cm}
\caption{Chinese Whisper clusters for the target word `web' extracted from COHA corpus for the time periods 1880-1900 and 1990-2000.}
%It is evident that in 1990-2000 the word `web' has emerged with a new cluster (C23) which is having a `technology' related sense, absent from the earlier time-period. 

\label{fig:coha}

\end{figure}
\begin{table}[!tbh]
%\vspace{-0.2cm}
\footnotesize
   \caption{Example cases, from the training set for the experiment on detecting known shifts. Evaluation has been done by annotators}
   %to check whether the word has actually emerged with a new sense.}
%\vspace{-0.4cm}
\begin{center}
    \begin{tabular}{|M{1cm}|M{3.5cm}|M{2.8cm}|}
    \hline
\centering \bf{Word} & \centering \bf{`birth' cluster} & \bf{Manual Evaluation} \\ \hline
\centering \emph{caller} &  phone, message, operator, customer, $\ldots$ & \textbf{Yes}, communication system related sense \\ \hline 
\centering \emph{courier} &  transport, purchase, company, delivery, $\ldots$ & \textbf{Yes}, marketing related sense \\ \hline
\centering public &  student, economist, general, $\ldots$ & \textbf{No} \\ \hline
\centering richness &  joy, happiness, stress, $\ldots$ & \textbf{No}  \\ \hline
    \end{tabular}
\label{tab:examples_training}
    \end{center}
\end{table}

\begin{table}[!tbh]
\footnotesize
   \caption{Example cases from COHA corpus, having linguistic shifts as suggested by prior literature and correctly detected by our approach. The discriminative feature shows the network measure which has changed the most over time.}
   %\vspace{-0.4cm}
\begin{center}
    \begin{tabular}{|M{1.1cm}|M{3cm}|M{1.9cm}|}
    \hline
\bf{Word} &	\bf{`birth' cluster} & \bf{Discriminative feature} \\ \hline
\centering virus & weapon, system, aircraft $\ldots$ & $\Delta$(SSM) \\ \hline 
\centering cell &  network, satellite, phone, $\ldots$ & $\Delta$(SSM)\\\hline

\centering monitor &  computer, TV, screen, $\ldots$ & $\Delta$(ED)\\\hline
\centering axis &   missile, fire, satellite, $\ldots$ & $\Delta$(ED) \\\hline
\centering broadcast &  TV, cable, service, $\ldots$ & $\Delta$(APL) \\\hline
\centering check &   wage, donation, fee, $\ldots$ & $\Delta$(APL) \\\hline
\centering film &  show, concert, script, $\ldots$ & $\Delta$(ED) \\\hline
\centering focus &   concern, ambition, $\ldots$ & $\Delta$(APL)\\\hline
\centering major &  university, discipline, $\ldots$ & $\Delta$(APL) \\\hline
\centering program &  project, database, testing, 
$\ldots$ & $\Delta$(ED) \\\hline
\centering record &   tape, card, disc, copy $\ldots$ & $\Delta$(SSM)\\\hline
\centering web & Web, Internet, network  $\ldots$ & $\Delta$(ED)\\\hline
    \end{tabular}
\label{tab:kw}
    \end{center}
%\vspace{-0.5cm}
\end{table}

\noindent {\bf Experiment details and results:} We first construct distributional thesauri (DT) networks~\cite{riedl2013scaling} for the COHA corpus at two different time points, 1880-1900 and 1990-2000. We apply Chinese Whispers algorithm~\cite{biemann2011structure} to produce a set of clusters for each target word in the DT network. The Chinese Whispers clusters for the target word `web' are shown in Figure~\ref{fig:coha}. Note that we have reported only some of the representative words for each cluster. 
Each of the clusters represents a particular sense of the target. We now compare the sense clusters extracted across two different time points to obtain the suitable signals of sense change following the approach proposed in Mitra {\sl et al.} Mitra {\sl et al.}~\shortcite{mitra2014s}. After getting the novel sense clusters, we pick up 50 random samples, of which 25 cases are flagged as `true birth' and the rest 25 cases are flagged as `false birth' by manual evaluation. We use these 50 samples as our training set for classification using SVM. Some of the examples of this training set are presented in Table~\ref{tab:examples_training}. We ensure that none of the words in the list $L$ is present in the training set. Using this training set for our proposed SVM classifier, we are successfully able to detect \textbf{80\%} of the cases (12 out of 15) from the list $L$ as having novel sense. Table~\ref{tab:kw} presents all of these detected words along with the novel senses and the discriminative network feature. Our method is unable to detect three cases -`gay', `guy' and `bush'. For `gay', since there is no sense cluster  in the older time period with `gay' being a noun, cluster comparison does not even detect the `birth' cluster of `gay'. The `birth' sense clusters for `guy', `bush' in the new time period, as detected by split-join algorithm contain general terms like ``someone, anyone, men, woman, mother, son" and ``cloud, air, sky, sunlight" respectively. As the network around these words did not change much over time, our method found it difficult to detect.
 Note that even though COHA corpus is substantially smaller than the Google n-grams corpus, our approach produces promising result, showing the usability of the proposed method with not so large corpus as well.  
\section{Conclusion}
%\vspace{-0.2cm}
\label{sec:conclusion}
In this paper, we showed how complex network theory can help improving the performance of otherwise challenging task of novel sense detection. This is the first attempt to apply concepts borrowed from complex network theory to deal with the problem of novel sense detection. We demonstrated how the change in the network properties of the induced subgraphs from a sense cluster can be used to improve the precision of novel sense detection significantly. Manual evaluation over two different time point pairs shows that the proposed SVM classification approach boosts the precision values from 0.23-0.32 to 0.74-0.86. Finally, from the experiments on the COHA corpus, we have also shown that our approach can reliably detect the words known to have sense shifts. We have made the human annotated data used in our experiments publicly available which could be used as gold standard dataset to validate systems built for novel sense detection\footnote{\url{https://tinyurl.com/ycj6ahud}}. 

This framework of precise novel sense detection of a word can be used by lexicographers as well as historical 
linguistics to design new dictionaries as well as updating the existing sense repositories like WordNet where candidate new senses can be semi-automatically detected and included, thus greatly reducing the otherwise required manual effort. Computational methods based on large diachronic corpora are considered to have huge potential to put a light on interesting language evolution phenomenon which can be useful for etymologists as well. In future, we plan to apply our methodology to different genres of corpus, like social network data, several product or movie reviews data which are becoming increasingly popular source for opinion tracking, to identify short-term changes in word senses or usages. These analyses would also provide insights on the evolution of language in a short span of time. Apart from that, we plan to extend our work to detect the dying senses of words; the senses which were used in the older texts, but are not being used in newer time anymore. Our ultimate goal is to prepare a generalized framework for accurate detection of sense change across languages and investigate the triggering factors behind language evolution as well.  

\if{0}
\section{Introduction}

The following instructions are directed to authors of papers submitted
to NAACL-HLT 2018 or accepted for publication in its proceedings. All
authors are required to adhere to these specifications. Authors are
required to provide a Portable Document Format (PDF) version of their
papers. \textbf{The proceedings are designed for printing on A4
paper.}
\fi

%\section{General Instructions}
\if{0}
Manuscripts must be in two-column format.  Exceptions to the
two-column format include the title, authors' names and complete
addresses, which must be centered at the top of the first page, and
any full-width figures or tables (see the guidelines in
Subsection~\ref{ssec:first}). {\bf Type single-spaced.}  Start all
pages directly under the top margin. See the guidelines later
regarding formatting the first page.  The manuscript should be
printed single-sided and its length
should not exceed the maximum page limit described in Section~\ref{sec:length}.
Pages are numbered for  initial submission. However, {\bf do not number the pages in the camera-ready version}.

By uncommenting {\small\verb|\aclfinalcopy|} at the top of this 
 document, it will compile to produce an example of the camera-ready formatting; by leaving it commented out, the document will be anonymized for initial submission.  When you first create your submission on softconf, please fill in your submitted paper ID where {\small\verb|***|} appears in the {\small\verb|\def\aclpaperid{***}|} definition at the top.

The review process is double-blind, so do not include any author information (names, addresses) when submitting a paper for review.  
However, you should maintain space for names and addresses so that they will fit in the final (accepted) version.  The NAACL-HLT 2018 \LaTeX\ style will create a titlebox space of 2.5in for you when {\small\verb|\aclfinalcopy|} is commented out.  

The author list for submissions should include all (and only) individuals who made substantial contributions to the work presented. Each author listed on a submission to NAACL-HLT 2018 will be notified of submissions, revisions and the final decision. No authors may be added to or removed from submissions to NAACL-HLT 2018 after the submission deadline.

\subsection{The Ruler}
The NAACL-HLT 2018 style defines a printed ruler which should be presented in the
version submitted for review.  The ruler is provided in order that
reviewers may comment on particular lines in the paper without
circumlocution.  If you are preparing a document without the provided
style files, please arrange for an equivalent ruler to
appear on the final output pages.  The presence or absence of the ruler
should not change the appearance of any other content on the page.  The
camera ready copy should not contain a ruler. (\LaTeX\ users may uncomment the {\small\verb|\aclfinalcopy|} command in the document preamble.)  
Reviewers: note that the ruler measurements do not align well with
lines in the paper -- this turns out to be very difficult to do well
when the paper contains many figures and equations, and, when done,
looks ugly. In most cases one would expect that the approximate
location will be adequate, although you can also use fractional
references ({\em e.g.}, the first paragraph on this page ends at mark $108.5$).

\subsection{Electronically-available resources}

NAACL-HLT provides this description in \LaTeX2e{} ({\small\tt naaclhlt2018.tex}) and PDF
format ({\small\tt naaclhlt2018.pdf}), along with the \LaTeX2e{} style file used to
format it ({\small\tt naaclhlt2018.sty}) and an ACL bibliography style ({\small\tt acl\_natbib.bst})
and example bibliography ({\small\tt naaclhlt2018.bib}).
These files are all available at
{\small\tt http://naacl2018.org/downloads/ naaclhlt2018-latex.zip}. 
 We
strongly recommend the use of these style files, which have been
appropriately tailored for the NAACL-HLT 2018 proceedings.

\subsection{Format of Electronic Manuscript}
\label{sect:pdf}

For the production of the electronic manuscript you must use Adobe's
Portable Document Format (PDF). PDF files are usually produced from
\LaTeX\ using the \textit{pdflatex} command. If your version of
\LaTeX\ produces Postscript files, you can convert these into PDF
using \textit{ps2pdf} or \textit{dvipdf}. On Windows, you can also use
Adobe Distiller to generate PDF.

Please make sure that your PDF file includes all the necessary fonts
(especially tree diagrams, symbols, and fonts with Asian
characters). When you print or create the PDF file, there is usually
an option in your printer setup to include none, all or just
non-standard fonts.  Please make sure that you select the option of
including ALL the fonts. \textbf{Before sending it, test your PDF by
  printing it from a computer different from the one where it was
  created.} Moreover, some word processors may generate very large PDF
files, where each page is rendered as an image. Such images may
reproduce poorly. In this case, try alternative ways to obtain the
PDF. One way on some systems is to install a driver for a postscript
printer, send your document to the printer specifying ``Output to a
file'', then convert the file to PDF.

It is of utmost importance to specify the \textbf{A4 format} (21 cm
x 29.7 cm) when formatting the paper. When working with
{\tt dvips}, for instance, one should specify {\tt -t a4}.
Or using the command \verb|\special{papersize=210mm,297mm}| in the latex
preamble (directly below the \verb|\usepackage| commands). Then using 
{\tt dvipdf} and/or {\tt pdflatex} which would make it easier for some.

Print-outs of the PDF file on A4 paper should be identical to the
hardcopy version. If you cannot meet the above requirements about the
production of your electronic submission, please contact the
publication chairs as soon as possible.

\subsection{Layout}
\label{ssec:layout}

Format manuscripts two columns to a page, in the manner these
instructions are formatted. The exact dimensions for a page on A4
paper are:

\begin{itemize}
\item Left and right margins: 2.5 cm
\item Top margin: 2.5 cm
\item Bottom margin: 2.5 cm
\item Column width: 7.7 cm
\item Column height: 24.7 cm
\item Gap between columns: 0.6 cm
\end{itemize}

\noindent Papers should not be submitted on any other paper size.
 If you cannot meet the above requirements about the production of 
 your electronic submission, please contact the publication chairs 
 above as soon as possible.

\subsection{Fonts}

For reasons of uniformity, Adobe's {\bf Times Roman} font should be
used. In \LaTeX2e{} this is accomplished by putting

\begin{quote}
\begin{verbatim}
\usepackage{times}
\usepackage{latexsym}
\end{verbatim}
\end{quote}
in the preamble. If Times Roman is unavailable, use {\bf Computer
  Modern Roman} (\LaTeX2e{}'s default).  Note that the latter is about
  10\% less dense than Adobe's Times Roman font.

\begin{table}[t!]
\begin{center}
\begin{tabular}{|l|rl|}
\hline \bf Type of Text & \bf Font Size & \bf Style \\ \hline
paper title & 15 pt & bold \\
author names & 12 pt & bold \\
author affiliation & 12 pt & \\
the word ``Abstract'' & 12 pt & bold \\
section titles & 12 pt & bold \\
document text & 11 pt  &\\
captions & 10 pt & \\
abstract text & 10 pt & \\
bibliography & 10 pt & \\
footnotes & 9 pt & \\
\hline
\end{tabular}
\end{center}
\caption{\label{font-table} Font guide. }
\end{table}

\subsection{The First Page}
\label{ssec:first}

Center the title, author's name(s) and affiliation(s) across both
columns. Do not use footnotes for affiliations. Do not include the
paper ID number assigned during the submission process. Use the
two-column format only when you begin the abstract.

{\bf Title}: Place the title centered at the top of the first page, in
a 15-point bold font. (For a complete guide to font sizes and styles,
see Table~\ref{font-table}) Long titles should be typed on two lines
without a blank line intervening. Approximately, put the title at 2.5
cm from the top of the page, followed by a blank line, then the
author's names(s), and the affiliation on the following line. Do not
use only initials for given names (middle initials are allowed). Do
not format surnames in all capitals ({\em e.g.}, use ``Mitchell'' not
``MITCHELL'').  Do not format title and section headings in all
capitals as well except for proper names (such as ``BLEU'') that are
conventionally in all capitals.  The affiliation should contain the
author's complete address, and if possible, an electronic mail
address. Start the body of the first page 7.5 cm from the top of the
page.

The title, author names and addresses should be completely identical
to those entered to the electronical paper submission website in order
to maintain the consistency of author information among all
publications of the conference. If they are different, the publication
chairs may resolve the difference without consulting with you; so it
is in your own interest to double-check that the information is
consistent.

{\bf Abstract}: Type the abstract at the beginning of the first
column. The width of the abstract text should be smaller than the
width of the columns for the text in the body of the paper by about
0.6 cm on each side. Center the word {\bf Abstract} in a 12 point bold
font above the body of the abstract. The abstract should be a concise
summary of the general thesis and conclusions of the paper. It should
be no longer than 200 words. The abstract text should be in 10 point font.

{\bf Text}: Begin typing the main body of the text immediately after
the abstract, observing the two-column format as shown in

the present document. Do not include page numbers.

{\bf Indent}: Indent when starting a new paragraph, about 0.4 cm. Use 11 points for text and subsection headings, 12 points for section headings and 15 points for the title.

\begin{table}
\centering
\small
\begin{tabular}{cc}
\begin{tabular}{|l|l|}
\hline
{\bf Command} & {\bf Output}\\\hline
\verb|{\"a}| & {\"a} \\
\verb|{\^e}| & {\^e} \\
\verb|{\`i}| & {\`i} \\ 
\verb|{\.I}| & {\.I} \\ 
\verb|{\o}| & {\o} \\
\verb|{\'u}| & {\'u}  \\ 
\verb|{\aa}| & {\aa}  \\\hline
\end{tabular} & 
\begin{tabular}{|l|l|}
\hline
{\bf Command} & {\bf  Output}\\\hline
\verb|{\c c}| & {\c c} \\ 
\verb|{\u g}| & {\u g} \\ 
\verb|{\l}| & {\l} \\ 
\verb|{\~n}| & {\~n} \\ 
\verb|{\H o}| & {\H o} \\ 
\verb|{\v r}| & {\v r} \\ 
\verb|{\ss}| & {\ss} \\\hline
\end{tabular}
\end{tabular}
\caption{Example commands for accented characters, to be used in, {\em e.g.}, \BibTeX\ names.}\label{tab:accents}
\end{table}

\subsection{Sections}

{\bf Headings}: Type and label section and subsection headings in the
style shown on the present document.  Use numbered sections (Arabic
numerals) in order to facilitate cross references. Number subsections
with the section number and the subsection number separated by a dot,
in Arabic numerals.
Do not number subsubsections.

\begin{table*}[t!]
\centering
\begin{tabular}{lll}
  output & natbib & previous ACL style files\\
  \hline
  \citep{Gusfield:97} & \verb|\citep| & \verb|\cite| \\
  \citet{Gusfield:97} & \verb|\citet| & \verb|\newcite| \\
  \citeyearpar{Gusfield:97} & \verb|\citeyearpar| & \verb|\shortcite| \\
\end{tabular}
\caption{Citation commands supported by the style file.
  The citation style is based on the natbib package and
  supports all natbib citation commands.
  It also supports commands defined in previous ACL style files
  for compatibility.
  }
\end{table*}
\fi

%{\bf Citations}: Citations within the text appear in parentheses
%as~\cite{Gusfield:97} or, if the author's name appears in the text
%itself, as Gusfield~\shortcite{Gusfield:97}.
%Using the provided \LaTeX\ style, the former is accomplished using
%{\small\verb|\cite|} and the latter with {\small\verb|\shortcite|} or {\small\verb|\newcite|}. Collapse multiple citations as in~\cite{Gusfield:97,Aho:72}; this is accomplished with the provided style using commas within the {\small\verb|\cite|} command, {\em e.g.}, {\small\verb|\cite{Gusfield:97,Aho:72}|}. Append lowercase letters to the year in cases of ambiguities.  
 %Treat double authors as
%in~\cite{Aho:72}, but write as in~\cite{Chandra:81} when more than two
%authors are involved. Collapse multiple citations %as
%in~\cite{Gusfield:97,Aho:72}. Also refrain from using full citations
%as sentence constituents.
\if{0}
We suggest that instead of
\begin{quote}
  ``\cite{Gusfield:97} showed that ...''
\end{quote}
you use
\begin{quote}
``Gusfield \shortcite{Gusfield:97}   showed that ...''
\end{quote}

If you are using the provided \LaTeX{} and Bib\TeX{} style files, you
can use the command \verb|\citet| (cite in text)
to get ``author (year)'' citations.

If the Bib\TeX{} file contains DOI fields, the paper
title in the references section will appear as a hyperlink
to the DOI, using the hyperref \LaTeX{} package.
To disable the hyperref package, load the style file
with the \verb|nohyperref| option: \\{\small
\verb|\usepackage[nohyperref]{naaclhlt2018}|}

\textbf{Digital Object Identifiers}:  As part of our work to make ACL
materials more widely used and cited outside of our discipline, ACL
has registered as a CrossRef member, as a registrant of Digital Object
Identifiers (DOIs), the standard for registering permanent URNs for
referencing scholarly materials.  As of 2017, we are requiring all
camera-ready references to contain the appropriate DOIs (or as a
second resort, the hyperlinked ACL Anthology Identifier) to all cited
works.  Thus, please ensure that you use Bib\TeX\ records that contain
DOI or URLs for any of the ACL materials that you reference.
Appropriate records should be found for most materials in the current
ACL Anthology at \url{http://aclanthology.info/}.

As examples, we cite \cite{P16-1001} to show you how papers with a DOI

will appear in the bibliography.  We cite
\fi
%\cite{C14-1001} 
%to show how
%papers without a DOI but with an ACL Anthology Identifier will appear
%in the bibliography. 

\if{0}

As reviewing will be double-blind, the submitted version of the papers
should not include the authors' names and affiliations. Furthermore,
self-references that reveal the author's identity, {\em e.g.},
\begin{quote}
``We previously showed \cite{Gusfield:97} ...''  
\end{quote}
should be avoided. Instead, use citations such as 
\begin{quote}
``\citeauthor{Gusfield:97} \shortcite{Gusfield:97}
previously showed ... ''
\end{quote}

Any preliminary non-archival versions of submitted papers should be listed in the submission form but not in the review version of the paper. NAACL-HLT 2018 reviewers are generally aware that authors may present preliminary versions of their work in other venues, but will not be provided the list of previous presentations from the submission form.

\textbf{Please do not use anonymous citations} and do not include
 when submitting your papers. Papers that do not
conform to these requirements may be rejected without review.

\textbf{References}: Gather the full set of references together under
the heading {\bf References}; place the section before any Appendices,
unless they contain references. Arrange the references alphabetically
by first author, rather than by order of occurrence in the text.
Provide as complete a citation as possible, using a consistent format,
such as the one for {\em Computational Linguistics\/} or the one in the 
{\em Publication Manual of the American 
Psychological Association\/}~\cite{APA:83}. Use of full names for
authors rather than initials is preferred. A list of abbreviations
for common computer science journals can be found in the ACM 
{\em Computing Reviews\/}~\cite{ACM:83}.

The \LaTeX{} and Bib\TeX{} style files provided roughly fit the
American Psychological Association format, allowing regular citations, 
short citations and multiple citations as described above.

Submissions should accurately reference prior and related work, including code and data. If a piece of prior work appeared in multiple venues, the version that appeared in a refereed, archival venue should be referenced. If multiple versions of a piece of prior work exist, the one used by the authors should be referenced. Authors should not rely on automated citation indices to provide accurate references for prior and related work.

{\bf Appendices}: Appendices, if any, directly follow the text and the
references (but see above).  Letter them in sequence and provide an
informative title: {\bf Appendix A. Title of Appendix}.

\subsection{Footnotes}

{\bf Footnotes}: Put footnotes at the bottom of the page and use 9
point font. They may be numbered or referred to by asterisks or other
symbols.\footnote{This is how a footnote should appear.} Footnotes
should be separated from the text by a line.\footnote{Note the line
separating the footnotes from the text.}

\subsection{Graphics}

{\bf Illustrations}: Place figures, tables, and photographs in the
paper near where they are first discussed, rather than at the end, if
possible.  Wide illustrations may run across both columns.  Color
illustrations are discouraged, unless you have verified that  
they will be understandable when printed in black ink.

{\bf Captions}: Provide a caption for every illustration; number each one
sequentially in the form:  ``Figure 1. Caption of the Figure.'' ``Table 1.
Caption of the Table.''  Type the captions of the figures and 
tables below the body, using 11 point text.

\subsection{Accessibility}
\label{ssec:accessibility}

In an effort to accommodate people who are color-blind (as well as those printing
to paper), grayscale readability for all accepted papers will be
encouraged.  Color is not forbidden, but authors should ensure that
tables and figures do not rely solely on color to convey critical
distinctions. A simple criterion: All curves and points in your figures should be clearly distinguishable without color.

% Min: no longer used as of NAACL-HLT 2018, following ACL exec's decision to
% remove this extra workflow that was not executed much.
% BEGIN: remove
%% \section{XML conversion and supported \LaTeX\ packages}

%% Following ACL 2014 we will also we will attempt to automatically convert 
%% your \LaTeX\ source files to publish papers in machine-readable 
%% XML with semantic markup in the ACL Anthology, in addition to the 
%% traditional PDF format.  This will allow us to create, over the next 
%% few years, a growing corpus of scientific text for our own future research, 
%% and picks up on recent initiatives on converting ACL papers from earlier 
%% years to XML. 

%% We encourage you to submit a ZIP file of your \LaTeX\ sources along
%% with the camera-ready version of your paper. We will then convert them
%% to XML automatically, using the LaTeXML tool
%% (\url{http://dlmf.nist.gov/LaTeXML}). LaTeXML has \emph{bindings} for
%% a number of \LaTeX\ packages, including the NAACL-HLT 2018 stylefile. These
%% bindings allow LaTeXML to render the commands from these packages
%% correctly in XML. For best results, we encourage you to use the
%% packages that are officially supported by LaTeXML, listed at
%% \url{http://dlmf.nist.gov/LaTeXML/manual/included.bindings}
% END: remove
\section{Translation of non-English Terms}

It is also advised to supplement non-English characters and terms
with appropriate transliterations and/or translations
since not all readers understand all such characters and terms.
Inline transliteration or translation can be represented in
the order of: original-form transliteration ``translation''.

\section{Length of Submission}
\label{sec:length}

The NAACL-HLT 2018 main conference accepts submissions of long papers and
short papers.
 Long papers may consist of up to eight (8) pages of
content plus unlimited pages for references. Upon acceptance, final
versions of long papers will be given one additional page -- up to nine (9)
pages of content plus unlimited pages for references -- so that reviewers' comments
can be taken into account. Short papers may consist of up to four (4)
pages of content, plus unlimited pages for references. Upon
acceptance, short papers will be given five (5) pages in the
proceedings and unlimited pages for references. 

For both long and short papers, all illustrations and tables that are part
of the main text must be accommodated within these page limits, observing
the formatting instructions given in the present document. Supplementary
material in the form of appendices does not count towards the page limit; see appendix A for further information.

However, note that supplementary material should be supplementary
(rather than central) to the paper, and that reviewers may ignore
supplementary material when reviewing the paper (see Appendix
\ref{sec:supplemental}). Papers that do not conform to the specified
length and formatting requirements are subject to be rejected without
review.

Workshop chairs may have different rules for allowed length and
whether supplemental material is welcome. As always, the respective
call for papers is the authoritative source.

\section*{Acknowledgments}

The acknowledgments should go immediately before the references.  Do
not number the acknowledgments section. Do not include this section
when submitting your paper for review.
\fi

% include your own bib file like this:
%\bibliographystyle{acl}
%\bibliography{naaclhlt2018}
%\bibliography{naaclhlt2018}

%\bibliographystyle{acl_natbib}

\if{0}
\appendix

\section{Supplemental Material}
\label{sec:supplemental}
Submissions may include resources (software and/or data) used in in the work and described in the paper. Papers that are submitted with accompanying software and/or data may receive additional credit toward the overall evaluation score, and the potential impact of the software and data will be taken into account when making the acceptance/rejection decisions. Any accompanying software and/or data should include licenses and documentation of research review as appropriate.

NAACL-HLT 2018 also encourages the submission of supplementary material to report preprocessing decisions, model parameters, and other details necessary for the replication of the experiments reported in the paper. Seemingly small preprocessing decisions can sometimes make a large difference in performance, so it is crucial to record such decisions to precisely characterize state-of-the-art methods. 

Nonetheless, supplementary material should be supplementary (rather
than central) to the paper. {\bf Submissions that misuse the supplementary 
material may be rejected without review.}
Essentially, supplementary material may include explanations or details
of proofs or derivations that do not fit into the paper, lists of
features or feature templates, sample inputs and outputs for a system,
pseudo-code or source code, and data. (Source code and data should
be separate uploads, rather than part of the paper).

The paper should not rely on the supplementary material: while the paper
may refer to and cite the supplementary material and the supplementary material will be available to the
reviewers, they will not be asked to review the
supplementary material.

Appendices ({\em i.e.} supplementary material in the form of proofs, tables,
or pseudo-code) should come after the references, as shown here. Use
\verb|\appendix| before any appendix section to switch the section
numbering over to letters.

\section{Multiple Appendices}
\dots can be gotten by using more than one section. We hope you won't
need that.
\fi

\end{document}